\documentclass[10pt,twocolumn,letterpaper]{article}

\usepackage{cvpr}
\usepackage{times}
\usepackage{microtype}
\usepackage{epsfig}
\usepackage{graphicx}
\usepackage{amsmath}
\usepackage{amssymb}

\usepackage{siunitx}
\usepackage{multirow}
\usepackage{booktabs}
\usepackage[skip=4.5pt]{caption}
\usepackage[skip=3pt]{subcaption}
\usepackage{cite}
\usepackage{fixfoot}
\usepackage{enumitem}

\usepackage[pagebackref=true,breaklinks=true,letterpaper=true,colorlinks,bookmarks=false]{hyperref}

\cvprfinalcopy % *** Uncomment this line for the final submission

\def\cvprPaperID{6356} % *** Enter the CVPR Paper ID here

\usepackage[capitalise]{cleveref}
\crefname{section}{Sec.}{Sections}
\Crefname{section}{Section}{Sections}
\crefname{equation}{}{}  % no abbreviated Eq. in front of an equation
\crefformat{footnote}{#2\footnotemark[#1]#3}

\makeatletter
\renewcommand{\paragraph}{%
  \@startsection{paragraph}{4}%
  {\z@}{1.2ex \@plus .3ex \@minus .2ex}{-0.7em}%
  {\normalfont\normalsize\bfseries}%
}
\makeatother

\ifcvprfinal\fi
\begin{document}

%%%%%%%%% TITLE
\title{Learning Physics-guided Face Relighting under Directional Light}

\author{
Thomas Nestmeyer\thanks{The majority of this work was performed while the authors were at Facebook Reality Labs. Corresponding author: {\tt T.Nestmeyer@gmail.com}}\\
% Max Planck Institute for Intelligent Systems\\
% Max Planck Institute for Intelligent Systems, T\"ubingen, Germany\\
MPI for Intelligent Systems%\\
% MPI-IS\\
% {\tt\small T.Nestmeyer@gmail.com}
\and
Jean-Fran\c{c}ois Lalonde\\
Universit\'e Laval%\\
% Universit\'e Laval, Quebec City, Canada\\
% {\tt\small jflalonde@gel.ulaval.ca}
\and
Iain Matthews\\
Epic Games%\\
% Epic Games, Pittsburgh, USA\\
% {\tt\small iain.matthews@epicgames.com}
\and
Andreas Lehrmann{$^*$}\\
Borealis AI%\\
% Borealis AI, Vancouver, Canada\\
% {\tt\small andreas.lehrmann@gmail.com}
}

\maketitle
\thispagestyle{empty}

%%%%%%%%% ABSTRACT
\begin{abstract}
Relighting is an essential step in realistically transferring objects from a captured image into another environment. For example, authentic telepresence in Augmented Reality requires faces to be displayed and relit consistent with the observer's scene lighting.
We investigate end-to-end deep learning architectures that both \emph{de-light} and \emph{relight} an image of a human face. Our  model decomposes the input image into intrinsic components according to a diffuse physics-based image formation model. We enable non-diffuse effects including cast shadows and specular highlights by predicting a residual correction to the diffuse render.
To train and evaluate our model, we collected a portrait database of 21 subjects with various expressions and poses. Each sample is captured in a controlled light stage setup with 32 individual light sources. Our method creates precise and believable relighting results and generalizes to complex illumination conditions and challenging poses, including when the subject is not looking straight at the camera.

% Project page:
Supplementary material can be found on our project page
% \url{https://ps.is.tue.mpg.de/research_projects/face-relighting}
\url{https://lvsn.github.io/face-relighting}
\end{abstract}

%%%%%%%%% BODY TEXT
\section{Introduction}

% \paragraph{Motivate task}
In recent years Augmented Reality~(AR) has seen widespread interest across a variety of fields, including gaming, communication, and remote work.
For an AR experience to be immersive, the virtual objects inserted in the environment should match the lighting conditions of their observed surroundings, even though they were originally captured under different lighting.
This task, known as \emph{relighting}, has a long history in computer vision with many seminal works paving the way for modern AR technologies~\cite{land1971retinex,barrow1978recoveringIntrinsics,peers2007reflectanceTransfer,barron2015sirfs,sengupta2018sfsnet}.

% \paragraph{previous solutions}
%Any approach attempting to solve this relighting problem is faced with two critical architectural design choices: the definition of internal representations and the processes operating on these internal representations. Depending on the shape and meaning of these two elements, we can place each model somewhere along an axis describing the extent of its \emph{structural assumptions}. 

\begin{figure}[t]
\centering
\def\lightSrc{0004}
\def\lightDst{0005}
\begin{subfigure}{0.32\linewidth}
\includegraphics[width=\linewidth]{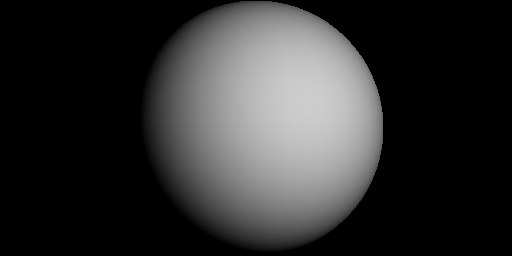}
\includegraphics[trim=0 0 0 50,clip,
% width=\linewidth]{figures/results/random_samples/150918-400004-\lightSrc-\lightDst_real_A_sRGB.png}
width=\linewidth]{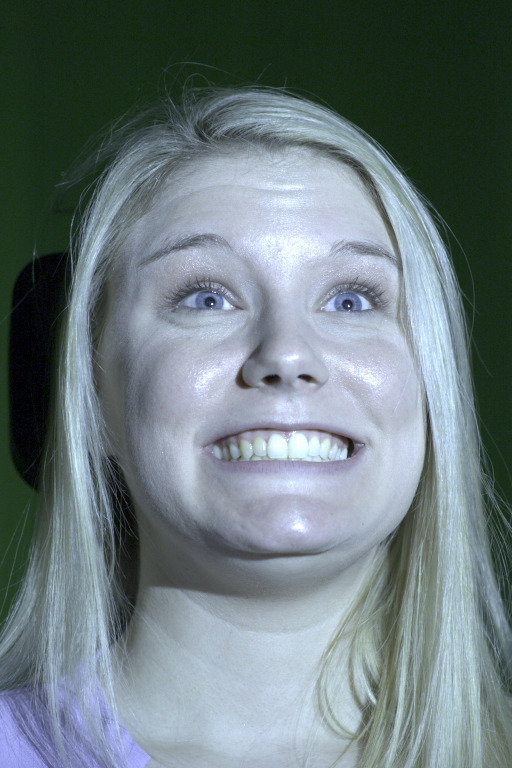}
\caption{$\ell_\text{src}$, input}
% \caption{$\ell_\text{src}$\\ input}
\end{subfigure}
\begin{subfigure}{0.32\linewidth}
\includegraphics[width=\linewidth]{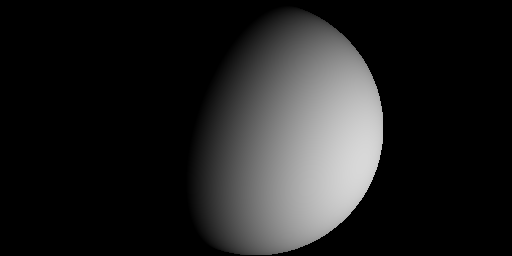}
\includegraphics[trim=0 0 0 50,clip,
% width=\linewidth]{figures/results/random_samples/150918-400004-\lightSrc-\lightDst_fake_B_sRGB.png}
width=\linewidth]{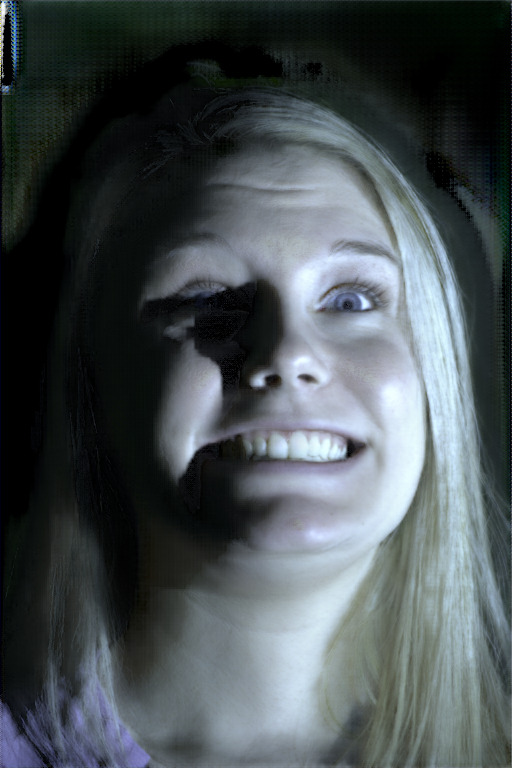}
\caption{$\ell_\text{dst}$, prediction}
\end{subfigure}
\begin{subfigure}{0.32\linewidth}
\includegraphics[width=\linewidth]{figures/lights/miniMugsy/256/\lightDst.png}
\includegraphics[trim=0 0 0 50,clip,
% width=\linewidth]{figures/results/random_samples/150918-400004-\lightSrc-\lightDst_real_B_sRGB.png}
width=\linewidth]{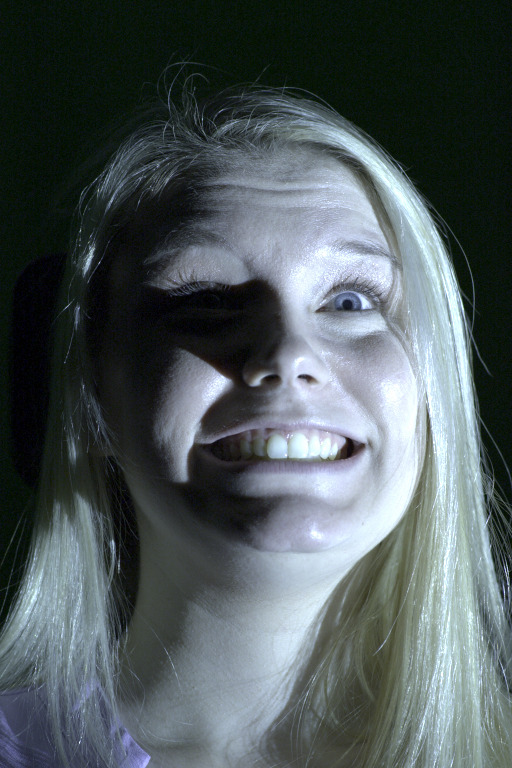}
\caption{$\ell_\text{dst}$, ground truth}
\end{subfigure}
\caption{{\bf Overview.} Given an unseen input image ({\bf a}) from the test set that was lit by the according directional light $\ell_\text{src}$ above it, we relight it towards the directional light $\ell_\text{dst}$ in ({\bf b}). To judge the performance, we provide the corresponding ground truth image in ({\bf c}).}
\label{fig:teaser}
\end{figure}

Relighting is often represented as a physics-based, two-stage process. First, \emph{de-light} the object in order to recover its intrinsic properties of reflectance, geometry, and lighting. Second, \emph{relight} the object according to a desired target lighting. This implies an exact instantiation of the rendering equation
% ~\cite{bartell1981bsdf} 
~\cite{Kajiya:1986} 
operating on lighting and surface reflectance representations capable of capturing the true nature of the light-material-geometry interactions. In practice, errors occur due to imperfect parametric models or assumptions. One common approximation is to assume diffuse materials~\cite{barron2015sirfs,sengupta2018sfsnet}.
Another approximation is smooth lighting, e.g.\ modeled as low-order spherical harmonics, which cannot produce hard shadows cast from point light sources like the sun.
We consider the hard problem of relighting human faces, which are known for both their complex reflectance properties including subsurface scattering, view-dependent and spatially-varying reflectance, but also for our perceptual sensitivity to inaccurate rendering.
Recent image-to-image translation approaches rely on deep learning architectures (e.g.~\cite{isola2017pix2pix}) that make no underlying structural assumption about the (re)lighting problem. Given enough representational capacity an end-to-end system can describe any underlying process, but is prone to large variance due to over-parameterization, and poor generalization due to physically implausible encodings. Test-time manipulation is also difficult with a semantically meaningless internal state. While this could potentially be alleviated with more training data, acquiring sufficient amounts is very time consuming. 

Recent approaches have demonstrated that explicitly integrating physical processes in neural architectures is beneficial in terms of both robust estimates from limited data and increased generalization~\cite{shu2017neuralFaceEditing,sengupta2018sfsnet,li2018learning}.
%due to the optimization process' coerced focus on a formation process rather than a data representation
However, these approaches have focused on the \emph{de-lighting} process, and use the simplified physical models for relighting that do not model non-diffuse effects such as cast shadows and specularities.

% \paragraph{Our solution}
In this work, we bridge the gap between the expressiveness of a physically unconstrained end-to-end approach and the robustness of a physics-based approach. In particular, we consider relighting as an image-to-image translation problem and divide the relighting task into two distinct stages:~a physics-based parametric rendering of estimated intrinsic components, and a physics-guided residual refinement. Our image formation model makes the assumption of directional light and diffuse materials.
The subsequent refinement process is \emph{conditioned} on the albedo, normals, and diffuse rendering, and dynamically accounts for shadows and any remaining non-diffuse phenomena.

% \subsection{Contributions}
We describe a neural architecture that combines the strengths of a physics-guided relighting process with the expressive representation of a deep neural network. Notably, our approach is end-to-end trained to simultaneously learn to both de-light \emph{and} relight.
We introduce a novel dataset of human faces under varying lighting conditions and poses, and demonstrate our approach can realistically relight complex non-diffuse materials like human faces.
Our directional lighting representation does not require assumptions of smooth lighting environments and allows us to generalize to arbitrarily complex output lighting as a simple sum of point lights.
To our knowledge, this is the first paper showing realistic relighting effects caused by strong directional lighting, such as sharp cast shadows, from a single input image. 

%!TEX root = 2020CVPR_Relighting.tex
\section{Related work}

\paragraph{Intrinsic images.}
Intrinsic image decomposition~\cite{barrow1978recoveringIntrinsics} and the related problem of shape from shading~\cite{zhang1999sfs} have inspired countless derived works. Of interest, 
%Barron and Malik~
\cite{barron2015sirfs} propose to simultaneously recover shape, illumination, reflectance and shading from a single image and rely on extensive priors to guide an inverse rendering optimization procedure. 
Other methods recover richer lighting representations in the form of environment maps given the known geometry~\cite{lombardi2016pami}. More recent approaches rely on deep learning for the same task, for example using a combination of CNN and guided/bilateral filtering~\cite{nestmeyer2017reflectanceFiltering} or a pure end-to-end CNN approach~\cite{fan2018revisitingDeepIntrinsics} with the common problem of hard to come by training data. Available datasets may include only sparse relative reflectance judgements~\cite{bell2014iiw}, or sparse shading annotations~\cite{Kovacs2017}, which limits learning and quantitative evaluation.

% \vspace{-1em}
% \paragraph{Relighting.} 
While many previous works focus on lighting estimation from objects~\cite{barron2015sirfs,lombardi2016pami,georgoulis2018reflectance,Meka:2018} or even entire images~\cite{karsch-tog-14,zhang-siga-16,gardner2017indoorIllumination,holdgeoffroy-cvpr-17,zhang2018pointLights}, few papers explicitly focus on the \emph{relighting} problem.
Notably, 
%Ren et al.~
\cite{ren2015imageBasedRelighting} use a small number of images as input, and, more recently, \cite{xu2018relightingFrom5lights} learn to determine which set of five light directions is optimal for relighting. Image-to-image translation~\cite{isola2017pix2pix} combined with novel multi-illumination datasets~\cite{murmann2019multiIlluminationDataset} has lately demonstrated promising results in full scene relighting. 
% When relighting from a single image, as we do in this work, \cite{cabral2011relightingTrees} propose a method specialized for tree canopies.

The explicit handling of moving hard shadows in
\cite{duchene2015multiviewIntrinsicsOutdoorsRelighting} and \cite{philip2019multiViewRelighting} is relevant. While both use multi-view inputs to relight outdoor scenes, our method works on a single input image to relight faces (our multi-view setup is only used to capture training data).
Similar to our work, \cite{yu2019inverseRendernet} regress to intrinsic components like albedo and normals, but their illumination model is spherical harmonics and therefore does not handle shadows.
\cite{neuralSengupta19} recently proposed a residual appearance renderer which bears similarities to our learned residual in that it models non-Lambertian effects. %, but do so with the goal of estimating the intrinsic components of indoor scenes while we focus on the relighting task.
Both of the latter works optimize for intrinsic decomposition, whereas we learn end-to-end relighting. Our intrinsic components are only used as a meaningful intermediate representation.

\paragraph{Face relighting.}
Lighting estimation from face images often focuses on normalization for improving face recognition. For example,~\cite{wen-cvpr-03} use spherical harmonics (SH) to relight a face image, and~\cite{wang-cvpr-07} use a Markov random field to handle sharp shadows not modeled by low-frequency SH models. Other face modeling methods have exploited approximate lighting estimates to reconstruct the geometry~\cite{lee-egsr-05,kemelmacher-eccv-2014} or texture~\cite{li-eccv-14}. In computer graphics similar ideas have been proposed for face replacement~\cite{bitouk-siggraph-08,dale-siggraph-asia-11}.
Low-frequency lighting estimation from a face has been explored in~\cite{shim2012faces,knorr-ismar-14,shahlaei-fg-15}. In contrast, \cite{nishino-siggraph-04} note that eyes reflect our surroundings and can be used to recover high frequency lighting. More closely related to our work,~\cite{calian2018faces2lightProbes} learn the space of outdoor lighting using a deep autoencoder and combine this latent space with an inverse optimization framework to estimate lighting from a face. However, their work is restricted to outdoor lighting and cannot be used explicitly for relighting. 

Of particular relevance to our work, neural face editing~\cite{shu2017neuralFaceEditing} and the related SfSNet~\cite{sengupta2018sfsnet} train CNNs to decompose a face image into surface normals, albedo, and SH lighting. These approaches also impose a loss on the intrinsic components, as well as a rendering loss which ensures that the \emph{combination} of these components is similar to the input image. 
FRADA~\cite{le2019frada} revisited the idea of relighting for improving face recognition with face-specific 3D morphable models~(similar to~\cite{shu2017neuralFaceEditing}), while we do not impose any face-specific templates.
Single image portrait relighting~\cite{relightingZhou19} bypasses the need for decomposition, while still estimating the illumination to allow editing. In a similar line of work, \cite{sun2019portraitRelighting} capture faces in a light stage using one light at a time, but then train using smoother illuminations from image based rendering which leads to artifacts when exposed to hard cast shadows or strong specularities.
Recently,~\cite{meka2019deepReflectanceFields} also used light stage data and train to relight to directional lighting as we do. However, their network expects a pair of images captured under spherical gradient illumination at test time, which can only be captured in a light stage. 
The portrait lighting transfer approach of~\cite{shu2018togPortraitLightingMassTransport} directly transfers illumination from a reference portrait to an input photograph to create high-quality relit images, but fails when adding/removing non-diffuse effects.

\begin{figure*}[t]
\centering
\begin{subfigure}{0.455\linewidth}
\includegraphics[width=\linewidth]{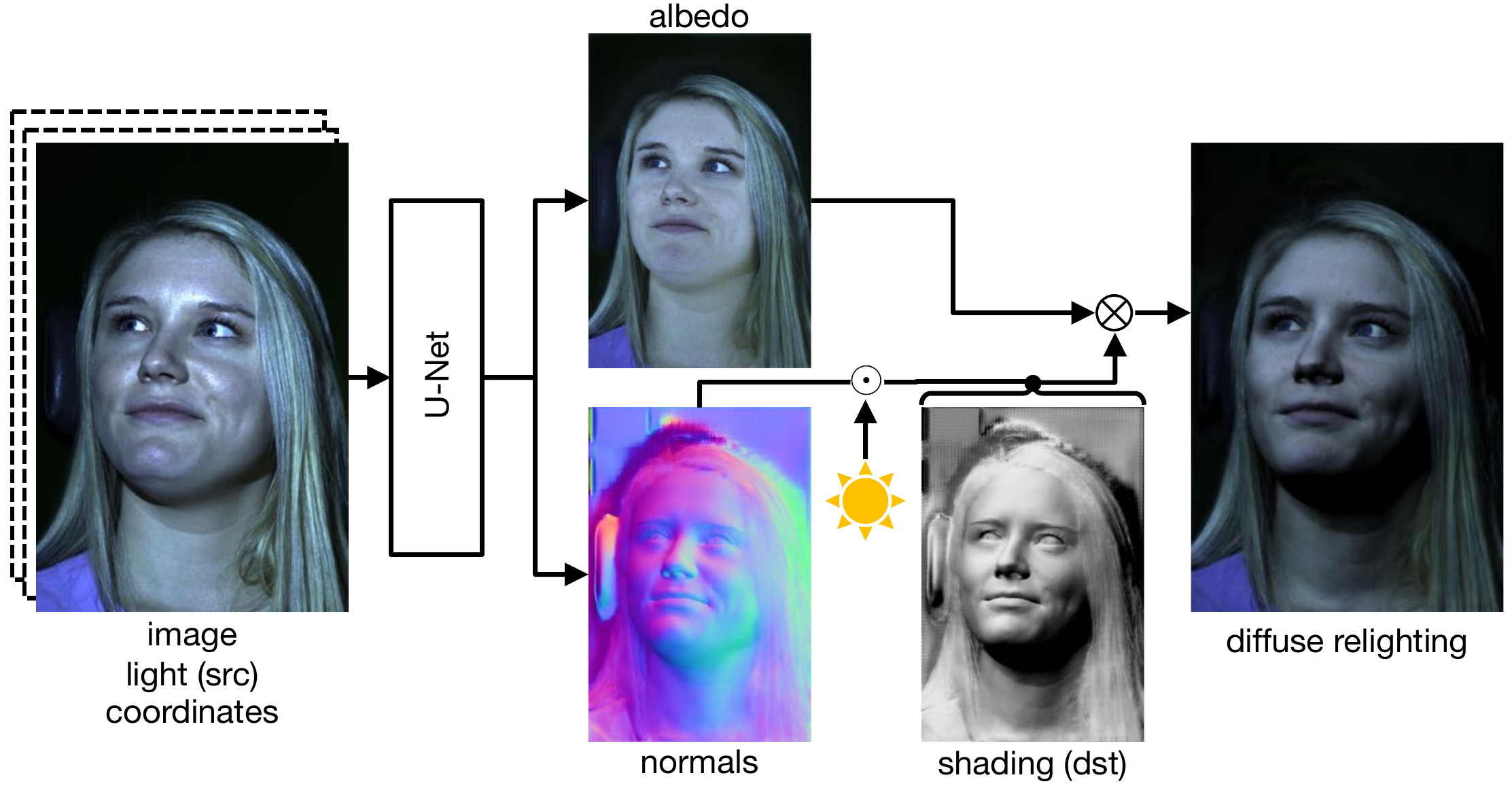}
\subcaption{Stage 1: Diffuse Rendering.}
\label{fig:stage1}
\end{subfigure}
\hfill
\begin{subfigure}{0.515\linewidth}
\includegraphics[width=\linewidth]{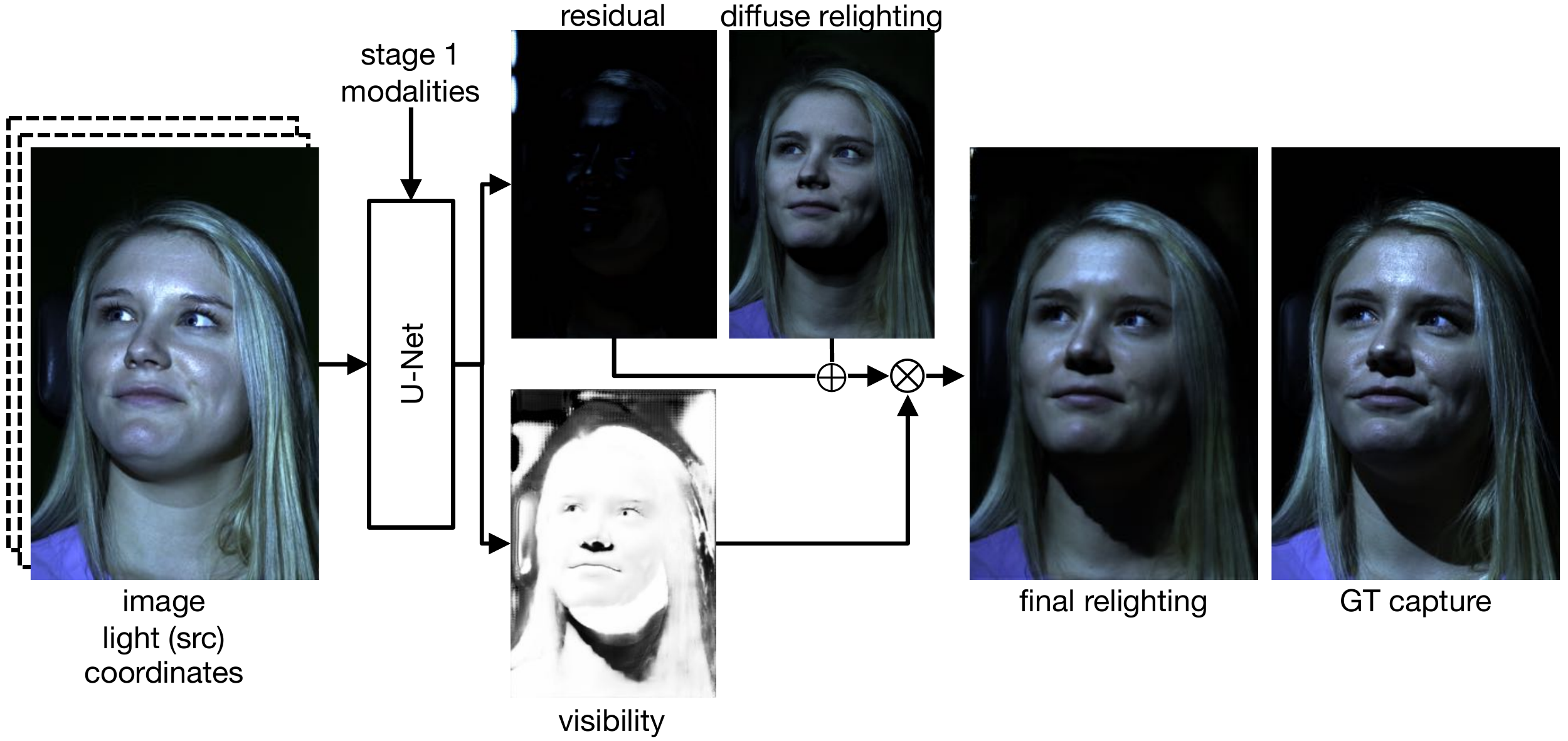}
\subcaption{Stage 2: Non-Diffuse Residual.}
\label{fig:stage2}
\end{subfigure}
\caption{{\bf Physics-guided relighting with structured generators.} Our generator consists of two stages modeling diffuse and non-diffuse effects. All intrinsic predictions are guided by losses w.r.t. photometric stereo reconstructions. {\bf (a)}~We use a U-Net with grouped convolutions to make independent predictions of the intrinsic components. Predicted normals are always re-normalized to unit vectors. Given a desired output lighting, we compute shading from normals and render a diffuse output. {\bf (b)}~Conditioned on all modalities inferred in~(\subref{fig:stage1}), we predict a non-diffuse residual and binary visibility map to model specularities, cast shadows, and other effects not captured by our instance of the rendering equation.}
\label{fig:architecture}
\end{figure*}

\section{Architecture}
\label{sec:architectureOverview}

The following two sections first introduce an image formation process~(\cref{sec:imageFormationModel}) and then describe its integration into a physics-based relighting architecture~(\cref{sec:architecture}).

\subsection{Image formation process}\label{sec:imageFormationModel}
The image formation process describes the physics-inspired operations transforming the intrinsic properties of a 3D surface to a rendered output. 
%While any such process can only be an approximation to reality, 
The majority of physics-based works are based on specific instantiations of the rendering equation~\cite{Kajiya:1986},
\begin{align}\label{eq:renderingEquation}
L_o(\omega_o) = \int_{\omega_i\in\Omega} f(\omega_i,\omega_o) L_i(\omega_i) \langle{\bf n},\omega_i\rangle\ \mathrm{d}\omega_i,
\end{align}
where $\omega_i,\omega_o$ are the incoming and outgoing light directions relative to the surface normal $\bf n$ at the surface point ${\bf X}_j$.
%Furthermore, 
$L_i(\omega_i)$ and $L_o(\omega_o)$ are the corresponding (ir)radiances, $f(\cdot,\cdot)$ is the BRDF describing the material's reflectance properties, and $\langle{\bf n},\omega_i\rangle$ is the attenuating factor due to Lambert's cosine law.

% \paragraph{Diffuse lighting.} 
This model is often simplified further by assuming a \emph{diffuse} decomposition into albedo $a \in \mathbb{R}$ and shading $s \in \mathbb{R}$,
\begin{align}\label{eq:diffuse}
a &= f(\omega_i,\omega_o),\quad\quad\quad\ \  [\textrm{const.}]\\
s &= \int_{\omega_i\in\Omega}L_i(\omega_i) \langle{\bf n},\omega_i\rangle\ \mathrm{d}\omega_i.
\end{align}
%Note that, while $a$ is constant for fixed ${\bf X}_j$, different image pixels ${\bf x}_j$ look at different 3D surface points ${\bf X}_j$, leading to spatially varying \mbox{albedo and shading maps.}

\paragraph{Non-diffuse effects.} A realistic relighting approach must relax modeling assumptions to allow complex reflectance properties such as subsurface scattering, transmission, polarization, etc., and if using~\cref{eq:diffuse} specularities. Unfortunately, learning a spatially varying BRDF model $f(\omega_i,\omega_o)$ based on a non-parametric representation is infeasible: assuming an image size of $512\times 768$ and a pixelwise discretization of the local half-angle space~\cite{matusik2003merl} would result in~\num{1.7e12} parameters. % 512*768*90*90*180*3
Learning a low-dimensional representation in terms of semantic parameters~\cite{burley12} seems like a viable alternative but is still prone to overfitting and cannot account for light-material-interactions outside of its parametric space.

We propose a hybrid approach and decompose $f$ into two principled components, a diffuse albedo $a$ and a light-varying residual $r$:
\begin{align}%\label{eq:ourBRDF}
f(\omega_i,\omega_o) &= a + r(\omega_i,\omega_o).
\end{align}
This turns~\cref{eq:renderingEquation} into
\begin{align}
% $
L_o(\omega_o) = as + \int_{\omega_i\in\Omega} r(\omega_i,\omega_o) L_i(\omega_i) \langle{\bf n},\omega_i\rangle\ \mathrm{d}\omega_i.
% $
\end{align}
For a light source with intensity $I(\omega_i)$, we can identify $L_i(\omega_i)=I(\omega_i) v(\omega_i)$, where $v\in\{0,1\}$ is the binary visibility of the light source.
Under the assumption of a single directional light source from $\widetilde{\omega}_i$, we integrate over one point only, so if we further write
$\widetilde{r}(\widetilde{\omega}_i,\omega_o)=r(\widetilde{\omega_i},\omega_o) I(\widetilde{\omega}_i) \langle {\bf n},\widetilde{\omega}_i\rangle$, we can re-formulate our rendering equation~\cref{eq:renderingEquation} to
\begin{align}\label{eq:ifp}
L_o({\omega_o}) = (as + \widetilde{r}(\widetilde{\omega_i},\omega_o))\cdot v({\widetilde{\omega}_i}).
\end{align}
This will be the underlying image formation process in all subsequent sections. While $as$ captures much of the diffuse energy across the image according to an explicit generative model, the residual $\widetilde{r}(\widetilde{\omega}_i,\omega_o)$ accounts for physical effects outside of the space representable by~\cref{eq:diffuse} and is modeled as a neural network (akin to \cite{neuralSengupta19}). We do not impose any assumptions on $r({\widetilde{\omega}_i,\omega_o})$, even allowing light subtraction, but do enforce $a$ to be close to the ground truth albedo of a diffuse model which we obtain from photometric stereo~\cite{xiong14pms}.% (see~\cref{sec:acquisition}).

% \vspace{-1em}
\paragraph{Discussion.} While directional lights are conceptually simple, they lead to challenging relighting problems. Our combination of an explicit diffuse rendering process and a non-diffuse residual (with implicit shading) serves several purposes: 
(1)~Describing most of the image intensities with a physics-based model means the output image will be more consistent with the laws of physics;
(2)~Describing specular highlights as residuals alleviates learning with a CNN;
(3)~Leaving the residual unconstrained (up to ground truth guidance) allows us to model effects that are not explainable by the BRDF, such as subsurface scattering and indirect light;
(4)~Modeling visibility explicitly helps, because the simple diffuse model does not handle cast shadows. At the same time, expecting the residual to take care of shadow removal by resynthesis is much harder than just masking it.

\subsection{Physics-guided relighting}\label{sec:architecture}
Presented with an input image $I_\text{src}$ that was lit by an input illumination $\ell_\text{src}$, our goal is to learn a generator~$G$, relighting $I_\text{src}$ according to a desired output illumination $\ell_\text{dst}$,
\begin{align}\label{eq:task}
G(I_\text{src},\ell_\text{src},\ell_\text{dst}) = I_\text{dst}.
\end{align}
At training time, we assume $\ell_\text{src}$ and $\ell_\text{dst}$ to be directional lights, which is known to be a particularly challenging instance of relighting and accurately matches our captured data~(see~\cref{sec:data}). At test time, this is not a limitation, since we can easily fit a set of directional lights to an environment map to perform more complex relighting (see~\cref{sec:extensions}).

Our physics-guided approach to solving the relighting task consists of a recognition model inferring intrinsic components from observed images (de-lighting) and a generative model producing relit images from intrinsic components (relighting). While the recognition model takes the form of a traditional CNN, the generative model follows our image formation process~(\cref{sec:imageFormationModel}) and is represented by structured layers with clear physical meaning. In line with~\cref{eq:ifp}, we implement the latter as a two-stage process: (Stage~1)~Using the desired target lighting, we compute shading from predicted normals and multiply the result with our albedo estimate to obtain a diffuse render; (Stage~2)~Conditioned on all intrinsic states predicted in stage 1, we infer a residual image and a visibility map, which we combine with the diffuse render according to~\cref{eq:ifp}. An illustration of this pipeline is shown in~\cref{fig:architecture}. Since all its operations are differentiable and directly stacked, this allows us to learn the proposed model in an end-to-end fashion from input to relit result.

We introduce losses for all internal predictions, i.e., albedo, normals, shading, diffuse rendering, visibility, and residual. We emphasize the importance of using the right loss function and refer to~\cref{sec:experiments:losses} for a comprehensive study. % regarding this choice. 
In order to obtain the corresponding guidance during training, we use standard photometric stereo reconstruction~\cite{xiong14pms}.

%!TEX root = 2020CVPR_Relighting.tex
\section{Data}\label{sec:data}
Our data comprises a diverse set of facial expressions captured under various lighting conditions.
% The following sections give a detailed overview of the acquisition~(\cref{sec:acquisition}) and augmentation~(\cref{sec:augmentation}) process.

\subsection{Acquisition}\label{sec:acquisition}
We record our data in a calibrated multi-view light-stage consisting of $6$ stationary Sony PMW-F55 camcorders and a total of $32$ white LED lights. The cameras record linear HDR images at $2048\!\times\!1080$ / \SI{60}{fps} and are synchronized with a Blackmagic sync generator that also triggers the LED lights. We flash one LED per frame and instruct our subjects to hold a static expression for the full duration of an LED cycle ($32$ frames $\sim$ \SI{0.53}{\second}).
In order to remove captures with motion, we filter our data based on the difference of two fully lit shots before/after each cycle. %we let the subjects trigger the recordings themselves while leaning their heads against a head rest. 
% Furthermore, 
For light calibration, we use a chrome sphere to recover directions and intensities in 3D~\cite{goldman2010lightCalibration} but subsequently express them with respect to each of the $6$ cameras, such that we obtain a total of $6\cdot32=192$ different light directions/intensities for each image.

We record a total of $482$ sequences from $21$ subjects, resulting in $482\cdot6\cdot32\cdot32 = 2{,}961{,}408$ relighting pairs. Each pair is formed using any one of the $32$ lights as input, and any one taken from the same sequence and same camera as output.
We split them into $81\%$ ($17$ subjects) for training, $9.5\%$ ($2$ subjects) for validation and $9.5\%$ ($2$ subjects) for testing.\footnote{The split into training, validation and testing was done manually in an effort to balance the demographics of the subjects.
% See~\cref{sec:appendix:datasetDetails} for details.} 
See supplementary material for details.} 
We did not ask the subjects to follow any specific protocol of facial expressions, besides being diverse, such that our evaluation on validation/test data is on \emph{both} unseen expressions and unseen subjects.

After extraction of the raw data, we use photometric stereo~(PMS) reconstruction~\cite{xiong14pms} to separate the input images $I$ into albedo $A$, shading $S$ with corresponding normals, and non-diffuse residual images $R=I-A \odot S$ per frame.

\subsection{Augmentation}\label{sec:augmentation}
Modern neural architectures are much better at interpolation than extrapolation.
It is therefore critical to cover the space of valid light transports as well as possible. To this end, we perform a series of data augmentations steps in an attempt to establish strong correlations throughout the parametric relighting space: (1)~We flip all training images along the horizontal and vertical axis, increasing the effective dataset size by a factor of $4$. Note that this also requires adaptation of the corresponding light directions and normals; (2)~We perform a linear scaling ${\bf x}' = s\cdot {\bf x}$, $s\sim\mathcal{U}_{[0.6,1.1]}$, of the images, shading, residuals and light intensities. In practice, we did not observe substantial benefits compared to training without scaling;
(3)~We randomly perturb the light calibration with Gaussian noise $n\sim\mathcal{N}(0,0.01^2)$ to improve generalization and account for minimal calibration errors; (4)~For quantitative results, we perform a spatial rescaling to $\frac{1}{8}$th of the original image resolution ($135\times256$), train on \emph{random crops} of size $128\times128$ and test on center crops with the same resolution to have comparability with \mbox{SfSNet}. Qualitative results are generated by rescaling to $\frac{1}{2}$ of the original resolution ($540\times1024$), trained on \emph{random crops} of size $512\times768$ and tested on center crops of that resolution.

\section{Experiments}
\label{sec:experiments}
Our models were implemented using PyTorch~\cite{paszke2017pytorch} with a U-Net~\cite{ronneberger2015unet} generator and PatchGAN~\cite{goodfellow2014gan} discriminator (for the final relit image) based on the implementations provided by pix2pix~\cite{isola2017pix2pix}.
%As commonly done in the decoder, transpose convolutions (sometimes referred to as Deconvolutions) are used. It is a known problem that this leads to a checkerboard pattern in the final resolution~\cite{odena2016deconvolutionCheckerboard}. We therefore also trained models with bilinear upsampling and following convolution instead, but got other artifacts instead. Our solution was to initialize the weights of ConvTranspose as $2\times2$ blocks with the same weight.
The images in our dataset are camera RAW, represented as 16-bit linear RGB values nominally in the range $[0,1]$. There is under- and over-shoot headroom, but for training and evaluation we clamp them into this range and linearly transform to $[-1,1]$ as input into the network. %Predicted normals are always renormalized to unit vectors.
% During training, losses are backpropagated from foreground (i.e.\ lit) pixels only.
% All results have been converted from linear to sRGB.

%%%%%%%%%%%%%%%%%%%%%%%%%%%%%%%%%%%%%%%%%%
\begin{table}%[t]
\caption{{\bf Loss selection.} We explore the influence of different training losses and evaluation metrics on direct image-to-image translation (``pix2pix'') and our structured guidance approach (``ours''). For each class, we show validation scores for all pairwise combinations of 5 training losses (rows) and the same 5 evaluation metrics (columns).
The best model for each evaluation metric is shown in bold.
% SSIM/MS-SSIM are expressed as dissimilarities~(\cref{eq:SSIM}).
}
\label{tab:eval:lossMatrix}
\centering
\vspace{-1pt}
\small
\scalebox{0.97}{\begin{tabular}{@{\hskip0pt}c@{\hskip8pt}p{1.55cm}>{\centering\arraybackslash}p{0.8cm}>{\centering\arraybackslash}p{0.8cm}>{\centering\arraybackslash}p{0.8cm}>{\centering\arraybackslash}p{0.8cm}>{\centering\arraybackslash}p{0.8cm}}
\toprule
\multirow{3}{*}{M} &  \multirow{3}{*}{\shortstack[l]{Training\\ Loss}} &\multicolumn{5}{c}{Evaluation Metric}\\
\cmidrule(lr){3-7}
&    & \multirow{2}{*}{$L_1$} & \multirow{2}{*}{$L_2$}   & \multirow{2}{*}{LPIPS} & \multirow{2}{*}{DSSIM} & MS-\\
&&&&&&DSSIM\\
\midrule

\multirow{5}{*}{\rotatebox{90}{pix2pix}}
& $L_1$     & .0452 & .0067 & .2564 & .1707 & .1144 \\
& $L_2$     & .0516 & .0082 & .2663 & .1911 & .1369 \\
& LPIPS  & .0424 & .0062 & \textbf{.1868} & .1440 & .0992 \\
& DSSIM      & \textbf{.0406} & \textbf{.0055} & .2138 & \textbf{.1378} & .0930 \\
& MS-DSSIM   & .0422 & .0058 & .2358 & .1547 & \textbf{.0913} \\

\midrule

% \multirow{5}{*}{\rotatebox{90}{partial}}
% & $L_1$     & .0366 & .0050 & .2312 & .1409 & .0861 \\
% & $L_2$     & .0410 & .0056 & .2641 & .1628 & .0968 \\
% & LPIPS  & .0363 & .0047 & \textbf{.1658} & .1282 & .0785 \\
% & SSIM      & \textbf{.0336} & \textbf{.0040} & .1962 & \textbf{.1159} & \textbf{.0689} \\
% & MS-SSIM   & .0366 & .0044 & .2257 & .1399 & .0758 \\

% \midrule

\multirow{5}{*}{\rotatebox{90}{ours}} 
& $L_1$   & .0406 & .0055 & .2237 & .1484 & .0913 \\
& $L_2$     & .0415 & .0056 & .2302 & .1547 & .0953 \\
& LPIPS  & {.0365} & {.0048} & \textbf{.1701} & {.1308} & {.0803} \\
& DSSIM      & \textbf{.0362} & \textbf{.0045} & .2008 & \textbf{.1270} & \textbf{.0793} \\
& MS-DSSIM   & .0410 & .0055 & .2165 & .1470 & .0910 \\

\bottomrule
\end{tabular}}
[M: model; LPIPS: \cite{zhang2018perceptual}; DSSIM: structured dissimilarity; \\ MS-DSSIM: multi-scale DSSIM]
\end{table}

\subsection{Evaluation metric}\label{sec:experiments:losses}

Quantitatively comparing the relit prediction~$\hat{I}_\text{dst}$ of the generator against the ground truth~$I_\text{dst}$ requires an appropriate error measure.
We consider the $L_1$ and $L_2$ norms but recognize that they do not coincide with human perceptual response. We also consider the ``Learned Perceptual Image Patch Similarity'' (LPIPS) loss suggested by~\cite{zhang2018perceptual} using the distance of CNN-features pretrained on ImageNet.
Another prevailing metric of image quality assessment is structural similarity (SSIM)~\cite{wang2004image} and its multi-scale variant (MS-SSIM)~\cite{wang2003msssim}.
In our evaluation, we use the corresponding dissimilarity measure
% \begin{align}
$\text{DSSIM}=\frac{1-\text{SSIM}}{2}$, %\label{eq:SSIM}
% \end{align}
and likewise for MS-SSIM, to consistently report errors.
%To provide a complete picture on our results, we report comparisons on all of those metrics.

When defining the loss function during training, the same choices of distance metrics are available. To densely evaluate their performance, we report in \cref{tab:eval:lossMatrix} the results of training all intrinsic layers with the same loss function from the options above.
Surprisingly, we conclude that, for our task, using DSSIM for the training loss consistently leads to models which generalize better on the validation set using most of the error metrics. The only exception is evaluation using the LPIPS metric, which is better when also trained using this metric.
Therefore, we chose the models trained on DSSIM for computing the final test results.

\def\tabsize{0.133\linewidth}
\def\tabwhite{0.15\linewidth}
\newcolumntype{C}[1]{>{\centering\let\newline\\\arraybackslash\hspace{0pt}}m{#1}}

\begin{figure*}%[t]
\def\imgsize{0.95\linewidth}
\centering
% \includegraphics[width=\imgsize]{figures/results/comparisons/150908-400004-0010-0025.jpg}\\
% \includegraphics[width=\imgsize]{figures/results/comparisons/150903-400004-0014-0008.jpg}\\
% 
% \includegraphics[width=\imgsize]{figures/results/comparisons/150902-400004-0010-0004.jpg}\\  % this is on arxiv
% \vspace{\negvspace}
% 
\includegraphics[trim=0 0 0 80,clip,
width=\imgsize]{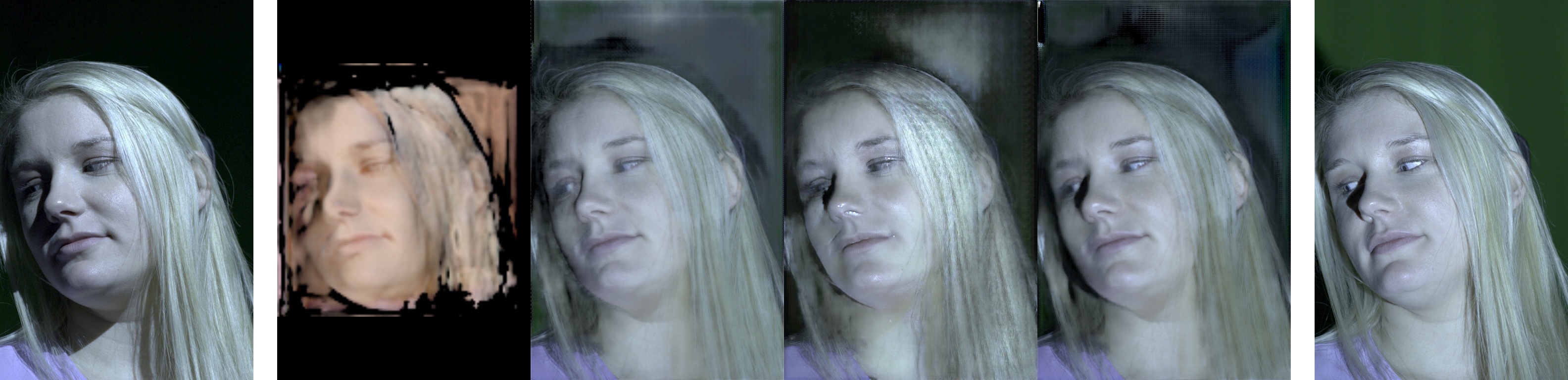}\\
\vspace{-3mm}
\includegraphics[width=\imgsize]{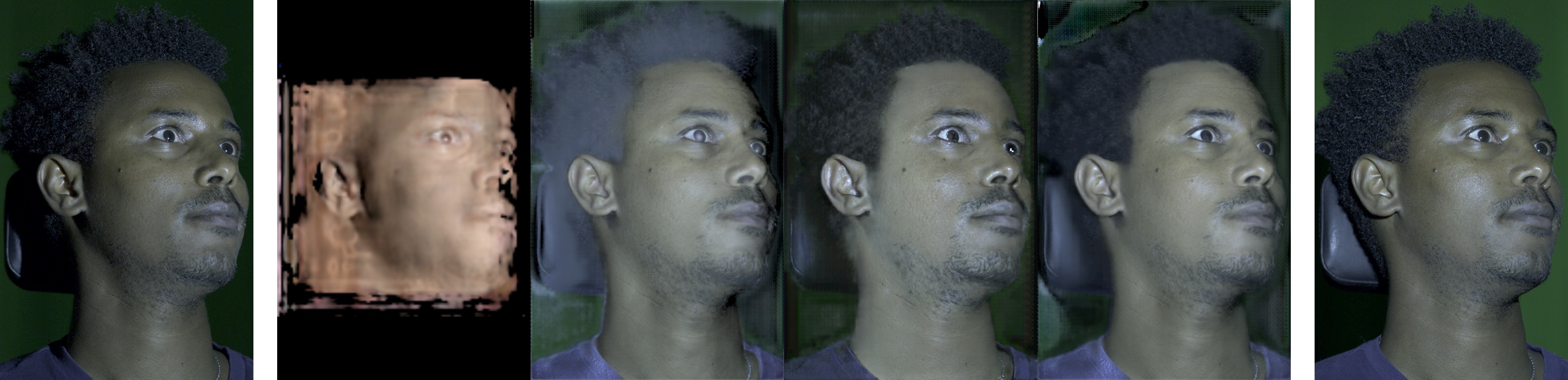}\\
\vspace{-5mm}
\includegraphics[trim=0 0 0 80,clip,
width=\imgsize]{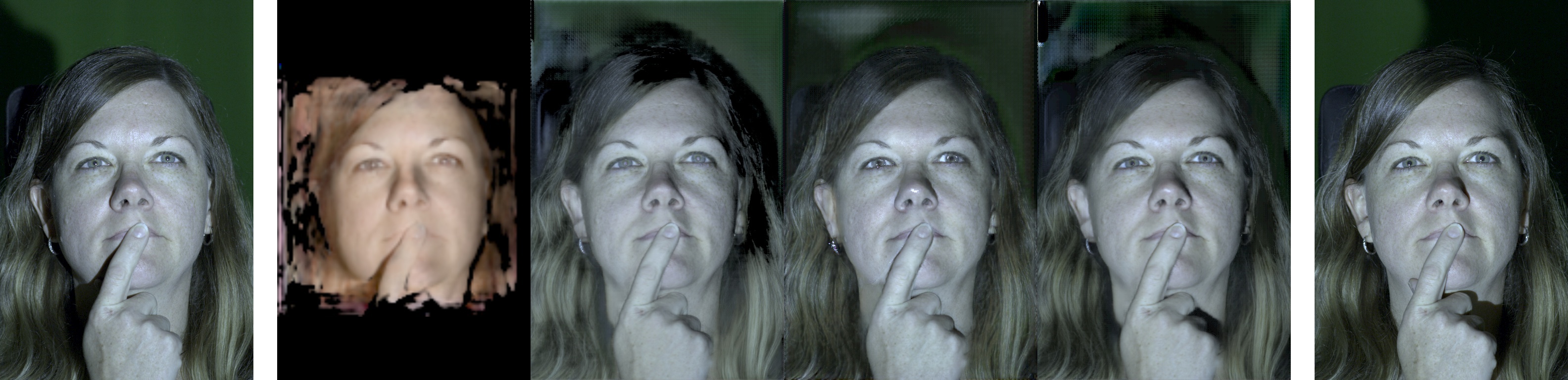}\\
\begin{tabular}{C{\tabwhite}C{\tabsize}C{\tabsize}C{\tabsize}C{\tabsize}C{\tabsize}C{\tabwhite}}
\small
(a)~input image &
\small
(b)~SfSNet~\cite{sengupta2018sfsnet} (pretrained) &
\small
(c)~SfSNet~\cite{sengupta2018sfsnet} (retrained) &
\small
(d)~pix2pix~\cite{isola2017pix2pix} &
\small
(e)~ours &
\small
(f)~ground truth
\end{tabular}
\caption[Qualitative evaluation on unseen subjects and expressions]{
{\bf Qualitative evaluation on unseen subjects and expressions.} We compare relighting ({\bf a})~the input image with ({\bf b}/{\bf c})~pretrained and retrained variants of SfSNet~\cite{sengupta2018sfsnet}, ({\bf d})~pix2pix~\cite{isola2017pix2pix}, and ({\bf e})~our model. In ({\bf f}), we show the ground truth capture of the given target illumination. Notice our model's ability to generate realistic shadows and specular highlights. %All results have been converted from linear to sRGB.
}
\label{fig:experiment:comparison}
\end{figure*}

%%%%%%%%%%%%%%%%%%%%%%%%%%%%%%%%%%%%%%%%%%
\subsection{Baseline comparisons}
\label{sec:experiments:comparison}

% We now provide quantitative~(\cref{tab:eval:testResults}) and qualitative~(\cref{fig:massTransfer,fig:experiment:comparison}) comparisons of our proposed architecture, to related work.
We now provide quantitative and qualitative comparisons of our proposed architecture, to related work.

\subsubsection{Baselines}\label{sec:baselines}

Our baselines comprise the following set of 
% traditional and neural architectures.
related methods.

\begin{itemize}[font=\bfseries, align=left, leftmargin=0em, labelwidth=-0.5em, itemsep=0.2em]%, noitemsep]
\item[PMS.] 
% \item{PMS.}
To understand the lower error bound of a diffuse model, we take albedo~$A$ and shading~$S$ from photometric stereo (PMS;~\cite{xiong14pms}) and diffuse render via $A\odot S$. We note that this model has access to all light configurations at the same time, with the desired target illumination amongst them. Since this gives an unfair advantage, we do not highlight results for this model in~\cref{tab:eval:testResults}.

\item[SfSNet (pretrained).] 
% \paragraph{SfSNet (pretrained).}
We take the pretrained %state-of-the-art 
network of SfSNet~\cite{sengupta2018sfsnet} and apply it to our data by using their decomposition into albedo and normals, but ignoring the output spherical harmonics estimate.
Instead, we compute target shading as the dot product of $\ell_\text{dst}$ and normals to have a direct comparison with our assumption of directional light and present the result after diffuse rendering.

\item[SfSNet (retrained).] 
% \paragraph{SfSNet (retrained).}
We retrain~\cite{sengupta2018sfsnet} on the calibrated PMS data and also provide the source illumination as input, to which our model has access as well. Compared to the pretrained model above, this baseline can be seen as a fairer comparison to SfSNet.

\item[Pix2pix.] 
% \paragraph{Pix2pix/no guidance.}
The arguably simplest way to learn~\cref{eq:task} from data is to instantiate $G$ as a traditional neural network consisting of a series of generic convolutional layers with no semantic meaning and no knowledge of the image formation process.%, such as the popular U-Net architecture~\cite{ronneberger2015unet}.
We adapt the pix2pix translation GAN~\cite{isola2017pix2pix} to our use case by conditioning the generator on the input image as well as the source and target illumination. This ensures an objective comparison with our more structured model, which also has access to lighting information.
\end{itemize}

\subsubsection{Evaluation}\label{sec:evaluation}

\paragraph{Qualitative evaluation.}
We compiled a collection of qualitative results in \cref{fig:experiment:comparison}.
%To align the low-resolution $128\times128$ pixel results obtained from SfSNet~\cite{sengupta2018sfsnet} with our predictions in a resolution of $512\times768$, we scale them to a width of $512$ pixels and center them vertically.
While the shading of \mbox{SfSNet} is smooth, it has a bias towards an albedo which probably resembles skin color in their training data and does not distinguish well between different skin tones and hair. As expected, retraining their model on our data leads to more accurate results. Still, due to the diffuse assumption, it looks flat compared to our more expressive model. It misses specularities and surface normals orthogonal to the light direction are missing ambient light from inter-reflections.
% The pretrained SfSNet model fails at more extreme head poses.
The pix2pix model generates promising results, but its domain-agnostic architecture often leads to physically implausible artifacts, such as missing shadows.
In comparison, the predictions of our proposed architecture are typically the most realistic, mainly due to its need to estimate a consistent albedo, as can be seen for example at the hair in the first row of~\cref{fig:experiment:comparison}.
The last row shows a hand occluding the face, leading to strong cast shadows that have to be introduced/removed. Our model using intrinsic guidance gracefully handles that case.

While our data allows foreground masking computed from PMS, we show the full image predictions for better judgment. At test time, an off-the-shelf face matting approach, e.g.~\cite{wadhwa2018mask}, could be used for cleaning the predictions.

We encourage the reader to look at more qualitative results of this type on our project page\footnote{\url{https://lvsn.github.io/face-relighting}}, where we also show relighting under a moving target illumination.

\paragraph{Quantitative evaluation.}
In~\cref{tab:eval:testResults} (first block), we analyze the quantitative performance of our model in the described scenario with known source illumination $\ell_\text{src}$. The test set comparison with SfSNet and the diffuse pix2pix baseline confirms the importance of our physics-based guidance and non-Lambertian residuals. An extension of our model without the assumption of known source illumination (second block) will be discussed in~\cref{sec:NoSourceIllumination}. For reference, the PMS reconstruction, restricted to a diffuse model but computed from multiple images, is also shown. 

\begin{table}%[t]
\caption{{\bf Quantitative evaluation.} We show a quantitative comparison of our approach to baseline methods. Performance on the test set is reported under the assumption of both known (`with') and unknown (`w/o') source illumination. All models have been trained with the DSSIM loss.
%SSIM/MS-SSIM are expressed as dissimilarities~(\cref{eq:SSIM}).
%The best model for each error metric is given in bold (excluding PMS with access to more data, see~\cref{sec:experiments:comparison}).
}
\label{tab:eval:testResults}
\centering
\vspace{-1pt}
\small
\scalebox{0.95}{\begin{tabular}{@{\hskip0pt}c@{\hskip6pt}p{2.5cm}@{\hskip4pt}>{\centering\arraybackslash}p{0.75cm}>{\centering\arraybackslash}p{0.75cm}>{\centering\arraybackslash}p{0.75cm}>{\centering\arraybackslash}p{0.75cm}>{\centering\arraybackslash}p{0.75cm}}
\toprule
\multirow{3}{*}{L} &  \multirow{3}{*}{\shortstack[l]{Model}} &\multicolumn{5}{c}{Evaluation Metric}\\
\cmidrule(lr){3-7}
&    & \multirow{2}{*}{$L_1$} & \multirow{2}{*}{$L_2$}   & \multirow{2}{*}{LPIPS} & \multirow{2}{*}{DSSIM} & MS-\\
&&&&&&DSSIM\\
\midrule

\multirow{3}{*}{\rotatebox{90}{with}}
& SfSNet (R)    & .0636 & \textbf{.0121} & .2508 & .1840 & .1277 \\  % test: (0.0636341925453325, 0.012136413815399726, 0.25097305019771365, 0.18399167053180698, 0.1276576417948191)
& pix2pix    & .0668 & .0144 & .2430 & .1832 & .1328 \\
% &partial guidance & \textbf{.0590} & \textbf{.0118} & .2195 & \textbf{.1609} & \textbf{.1111} \\
& ours   & \textbf{.0609} & .0123 & \textbf{.2144} & \textbf{.1618} & \textbf{.1138} \\  % test: (0.06091940295284205, 0.012344692507463163, 0.21444311119108903, 0.16176077496700483, 0.11378403123575154)

\midrule

\multirow{3}{*}{\rotatebox{90}{w/o}}
& SfSNet (P)    & .1359 & .0424 & .4703 & .3221 & .3121 \\ % test: (0.13594753022795303, 0.042411873077250216, 0.470286029890873, 0.3220551056666006, 0.31207697021723885)
%& SfSNet (R)    & . & . & . & . & . \\  % test: 
& pix2pix    & .0815 & .0189 & .2783 & .2076 & .1623 \\  % test: (0.08148040715088961, 0.01889105280801269, 0.2782719389710874, 0.2076321785737675, 0.1622526306866807) validation: & .0517 & .0082 & .2487 & .1735 & .1257 \\
% &partial guidance& .0695 & .0144 & \textbf{.2325} & .1801 & .1353 \\  % test: (0.06954093688489309, 0.014439117496509953, 0.232520477337562, 0.1801113413581962, 0.13530873967241924) validation & .0475 & .0070 & .2235 & .1535 & .1132 \\
& ours & \textbf{.0684} & \textbf{.0142} & \textbf{.2273} & \textbf{.1763} & \textbf{.1316} \\  % test: (0.0683754534445653, 0.01418593971994223, 0.22725898703377378, 0.1762899003148777, 0.1315923688193769) validation: & .0446 & .0064 & .2203 & .1467 & .1018 \\

\midrule
&PMS        & .0391 & .0047 & .1630 & .1125 & .0561 \\  % test: (0.03910422243739627, 0.004728864346737025, 0.16296064917381423, 0.11251663215031492, 0.056071903796095446)

\bottomrule
\end{tabular}}
{[L: access to source illumination; LPIPS: \cite{zhang2018perceptual}; \\ DSSIM: structured dissimilarity, MS-DSSIM: multi-scale DSSIM;  \\ PMS: photometric stereo; P/R: (p)retrained]}
\end{table}

\subsection{Additional qualitative comparisons}
We provide more qualitative comparisons to the following recent portrait relighting approaches in~\cref{fig:massTransfer}.
%which differ in their light representation:
% % \begin{itemize}[font=\bfseries, align=left, leftmargin=0em, labelwidth=-1em, noitemsep]
% \begin{itemize}[font=\bfseries, align=left, leftmargin=0em, labelwidth=-0.5em, itemsep=0.2em]%, noitemsep]

% \item[Mass transport relighting.]
% \paragraph{Mass transport relighting~\cite{shu2018togPortraitLightingMassTransport}.}
The \emph{mass transport relighting}~\cite{shu2018togPortraitLightingMassTransport} approach is different in that it defines the target lighting as that of \emph{another} portrait. To match those conditions, we set the desired output to directly be the ground truth reference image. Despite these optimal conditions, \cite{shu2018togPortraitLightingMassTransport} fails to generate specular highlights and cast shadows, which are well-captured by our technique.

\emph{Single image portrait relighting} using an environment map is learned in~\cite{sun2019portraitRelighting}. Training images are produced by compositing multiple `one light at a time' captures. As already discussed in~\cite{sun2019portraitRelighting}, the method fails on strong light.

% \item[Deep portrait relighting.] 
% \paragraph{Deep portrait relighting~\cite{relightingZhou19}.}
Finally,~\cite{relightingZhou19} learn to do \emph{deep portrait relighting} using a spherical harmonics representation, which also handles smooth lighting exclusively, as can be seen in~\cref{fig:massTransfer}.

% \end{itemize}

% %%%%%%%%%%%%%%%%%%% more qualitative comparison
\def\imgsize{0.1985\linewidth}
\begin{figure}%[t]
\setlength{\lineskip}{0pt}
% \def\whitespace{0.015\columnwidth}
% trim notation: left bottom right top
% trim=0 80 0 60,clip,  % quite a lot trimmed
% trim=0 40 0 40,clip,  % a bit more of the image
\centering
\includegraphics[trim=0 40 0 40,clip,width=\imgsize]
% {figures/comparison_related/mass_transfer/150901_400004_0003.png}%
{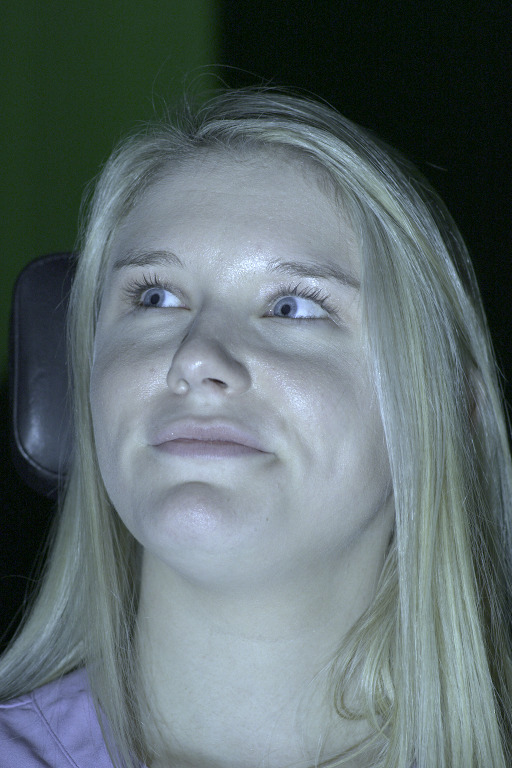}%
\hfill
\includegraphics[trim=0 40 0 40,clip,width=\imgsize]
% {figures/comparison_related/mass_transfer/{result__tgt_150901_400004_0003.png_ref_150901_400004_0010}.png}%
{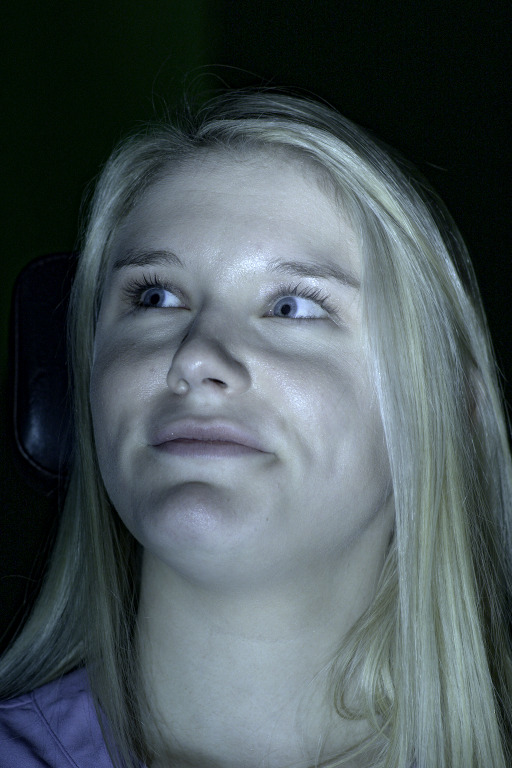}%
\includegraphics[trim=0 40 0 40,clip,width=\imgsize]{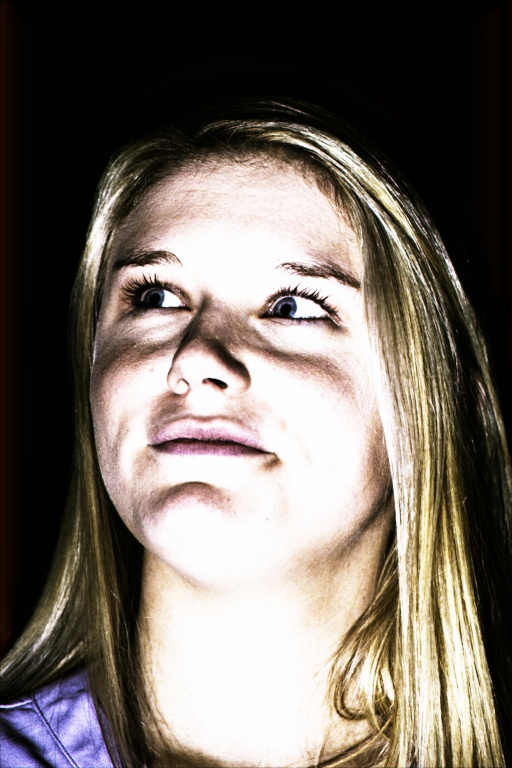}%
\includegraphics[trim=0 40 0 40,clip,width=\imgsize]{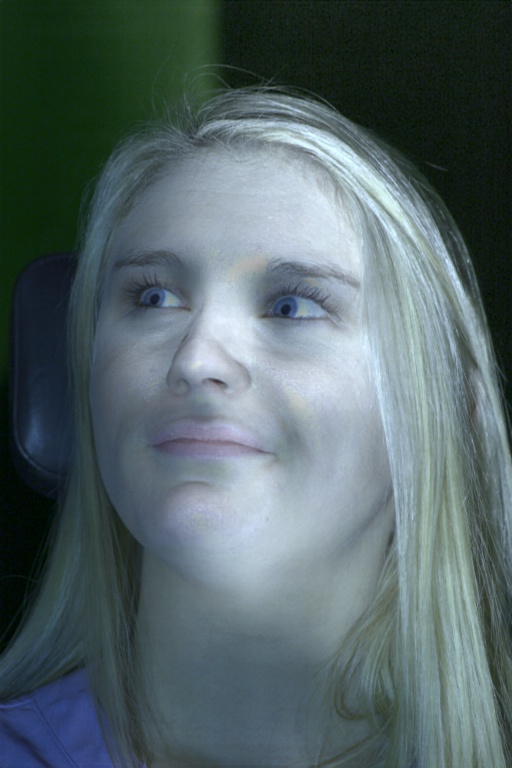}%
\includegraphics[trim=0 40 0 40,clip,width=\imgsize] 
% {figures/comparison_related/mass_transfer/150901-400004-0003-0010_fake_B_sRGB.png}
{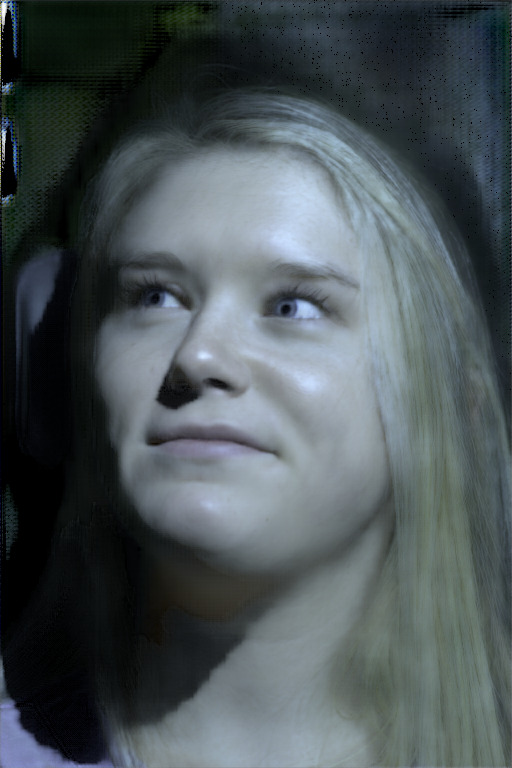}
\\
% second row
\includegraphics[trim=0 40 0 40,clip,width=\imgsize]
%{figures/comparison_related/mass_transfer/150901_400004_0010.png}%
{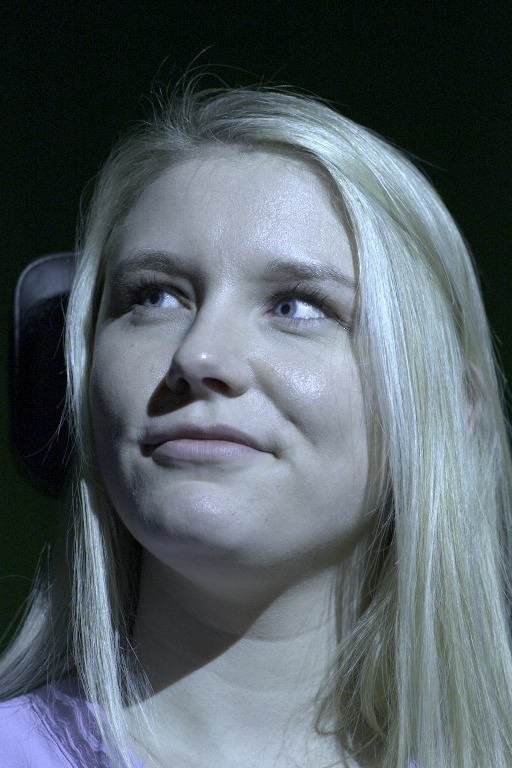}%
% DPR
\hfill
% Mass Transfer
\includegraphics[trim=0 40 0 40,clip,width=\imgsize]
% {figures/comparison_related/mass_transfer/{result__tgt_150901_400004_0010.png_ref_150901_400004_0003}.png}%
{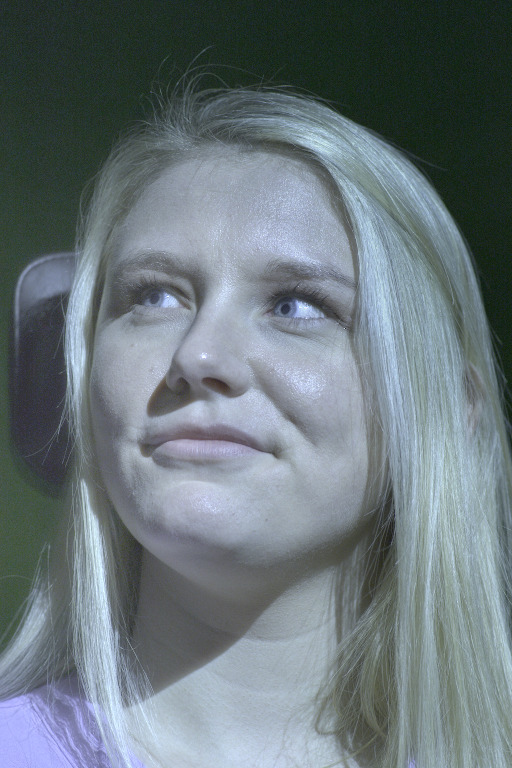}%
%\caption{MT~\cite{shu2018togPortraitLightingMassTransport}}
\includegraphics[trim=0 40 0 40,clip,width=\imgsize]{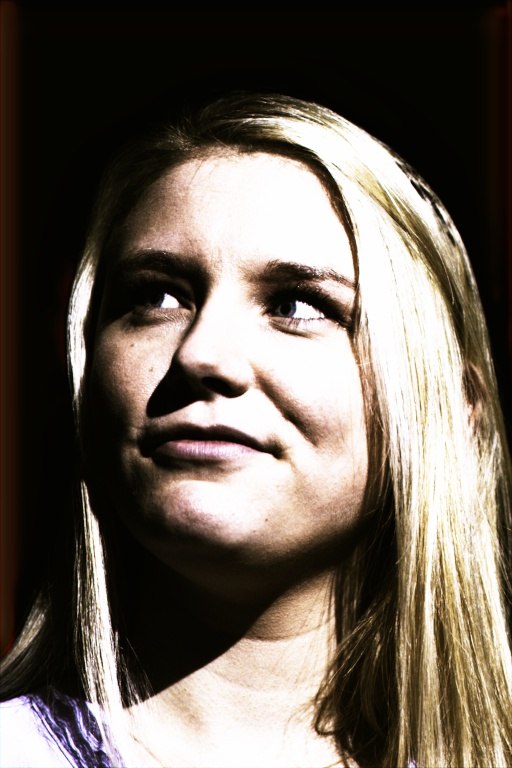}%
\includegraphics[trim=0 40 0 40,clip,width=\imgsize]{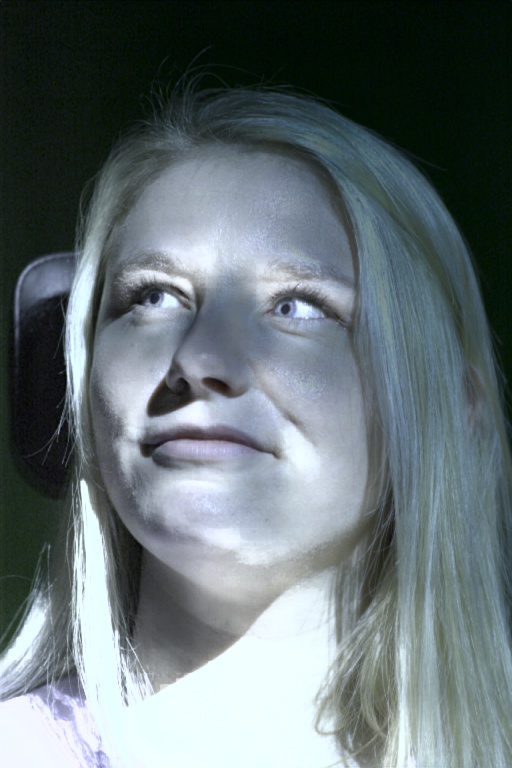}%
%\caption{DPR~\cite{relightingZhou19}}
% ours
\includegraphics[trim=0 40 0 40,clip,width=\imgsize]
% {figures/comparison_related/mass_transfer/150901-400004-0010-0003_fake_B_sRGB.png}%
{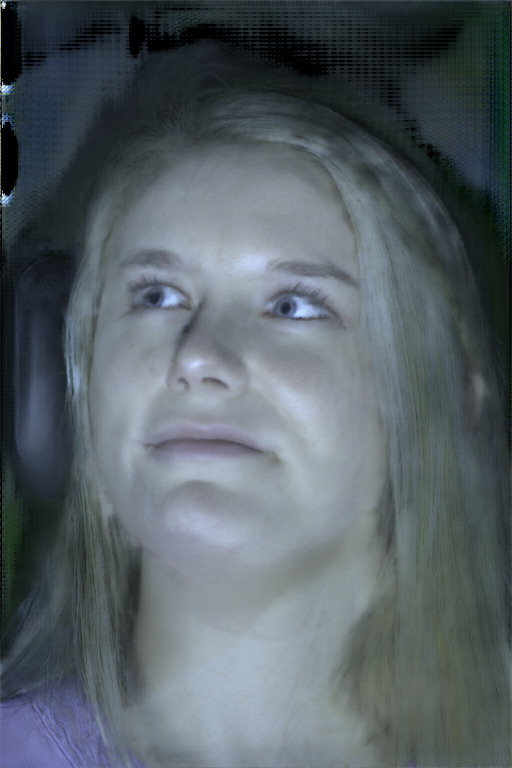}%
%\caption{ours}

\begin{tabular}{@{}C{\imgsize}@{}C{\imgsize}@{}C{\imgsize}@{}C{\imgsize}@{}C{\imgsize}@{}}
(a) input & (b) MT & (c) SIPR & (d) DPR & (e) ours
\end{tabular}

\caption{{\bf Additional qualitative comparisons.}
We relight the input in the first row to the input in the second row, and vice versa. Results in the Mass Transport~(MT)~\cite{shu2018togPortraitLightingMassTransport} approach and Single Image Portrait Relighting~(SIPR)~\cite{sun2019portraitRelighting} were provided by the authors. For Deep Portrait Relighting~(DPR)~\cite{relightingZhou19}, we use their provided code and approximate the light directions manually.
}
\label{fig:massTransfer}
\end{figure}

%%%%%%%%%%%%%%%%%%%%%%%%%%%%%%%%%%%%%%%%%% Extensions
\section{Extensions}
\label{sec:extensions}
We now demonstrate that our model successfully generalizes to different scenarios, including unknown input illumination, relighting with environment maps, and relighting images captured in the wild. 

%%%%%%%%%%%%%%% no source illumination
\subsection{Relighting with unknown source illumination}\label{sec:NoSourceIllumination}
% All our models made use of the source and target illumination. %~(see~\cref{sec:viaNormals}). 
While we cannot remove the need for the target illumination, information about the source illumination is already contained in the input image, allowing for implicit learning of $\ell_\text{src}$. To illustrate our model's ability to extract these signals, we trained a version of our architecture without explicit access to the input lighting; these results, as well as a comparison to the corresponding baseline variants, are shown in~\cref{tab:eval:testResults} (second block) and~\cref{fig:experiments:sourceIllum}. As expected, all models incur a small drop in performance compared to their counterparts with explicit knowledge. Nonetheless, our model \emph{without} access to the source illumination achieves similar (and in some cases better) performance than the pix2pix model \emph{with} access to the source illumination.

\def\imgsize{0.2475\linewidth}
\begin{figure}%[t]
\centering
\includegraphics[width=\imgsize]{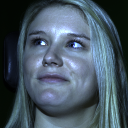}%
\hspace{0mm}
\includegraphics[width=\imgsize]{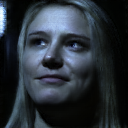}%
\includegraphics[width=\imgsize]{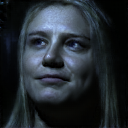}%
\includegraphics[width=\imgsize]{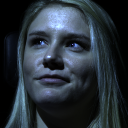}\\
\begin{tabular}{@{}C{\imgsize}@{}C{\imgsize}@{}C{\imgsize}@{}C{\imgsize}@{}}
(a) input & (b) with & (c) w/o & (d) GT 
\end{tabular}

\caption{
{\bf Relighting with unknown source illumination.}
({\bf a})~Input image.
% ({\bf b})~Relighting model \textit{with} access to source illumination.
% ({\bf c})~Relighting model \textit{without} access to source illumination.
({\bf b/c})~Our relighting model \textit{with} and \textit{without} access to source illumination.
({\bf d})~Ground truth output.
}
\label{fig:experiments:sourceIllum}
\end{figure}

%%%%%%%%%%%%%%%%%%%%%%%%%%%%%%%%%%%%%%%%%%%%%%%%%%%%%%%%%%%%%%%%%
% environment map relighting
\subsection{Relighting with environment maps}

Directional light sources are a very general representation, and our approach easily allows for relighting with environment maps. While more principled approaches like importance sampling are available~\cite{agarwal2003sampleEnvMap}, for illustrative purposes here we simply sample environment maps by downscaling them to $64\times32$ pixels and instantiating our relighting prediction with one light direction per pixel. Since light is additive, we then mix the resulting predictions according to their color and intensity.
Examples of a relit scene with three environment maps are shown in~\cref{fig:experiments:envMap}, more results are contained in the supplementary material on our project page.

\begin{figure}[t]
\setlength{\lineskip}{0pt}
\def\envOne{uffizi-0000}
\def\envTwo{esplanade-0000}
\def\envThree{grace-0127}
\def\envFour{grace-0000}
\def\envFive{fire-0127}
\centering
%%%%% lights
\includegraphics[width=\imgsize]{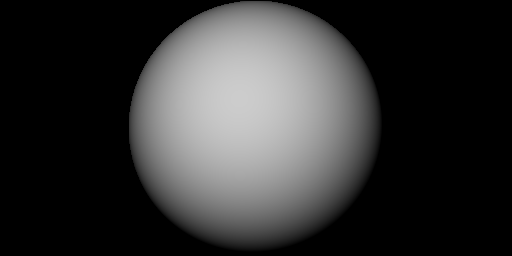}
\hfill
% PNG
% \includegraphics[width=\imgsize]{figures/lights/envmaps/\envOne_sRGB.png}%
% % \includegraphics[width=\imgsize]{figures/lights/envmaps/\envTwo_sRGB.png}%
% \includegraphics[width=\imgsize]{figures/lights/envmaps/\envThree_sRGB.png}%
% \includegraphics[width=\imgsize]{figures/lights/envmaps/\envFour_sRGB.png}%
% % \includegraphics[width=\imgsize]{figures/lights/envmaps/\envFive_sRGB.png}
% % \includegraphics[width=\imgsize]{figures/lights/envmaps/\envSix_sRGB.png}
% JPG
\includegraphics[width=\imgsize]{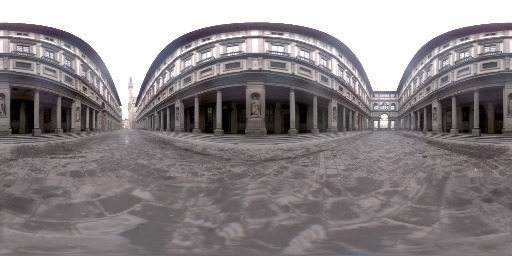}%
\includegraphics[width=\imgsize]{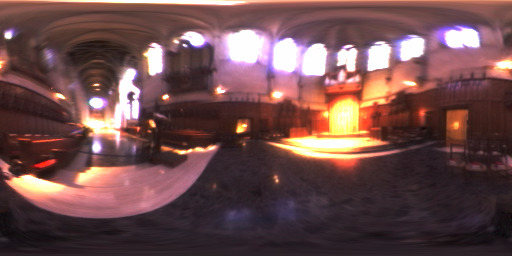}%
\includegraphics[width=\imgsize]{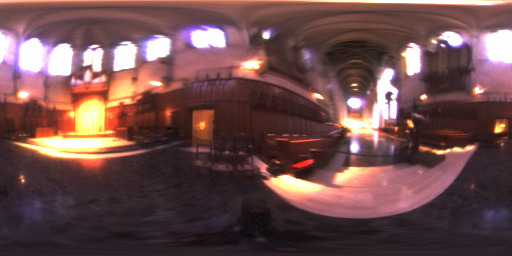}%
\\
\def\subject{150906}
\includegraphics[trim=0 11 0 64,clip,width=\imgsize] {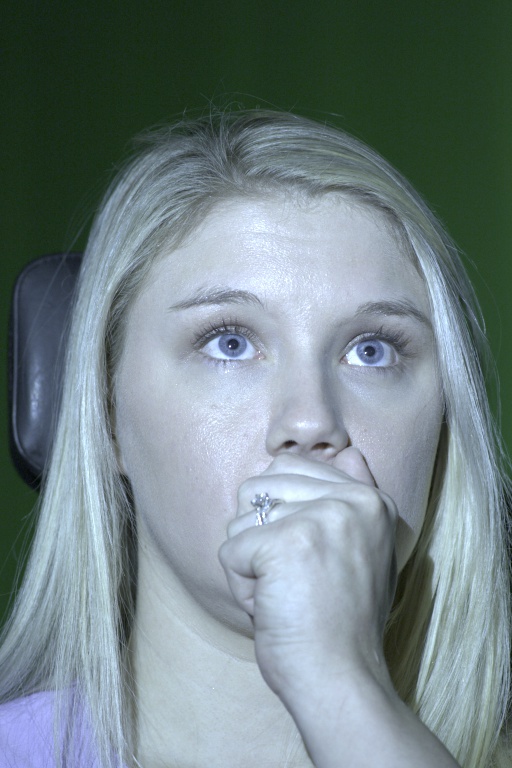}
\includegraphics[trim=0 11 0 64,clip,width=\imgsize] {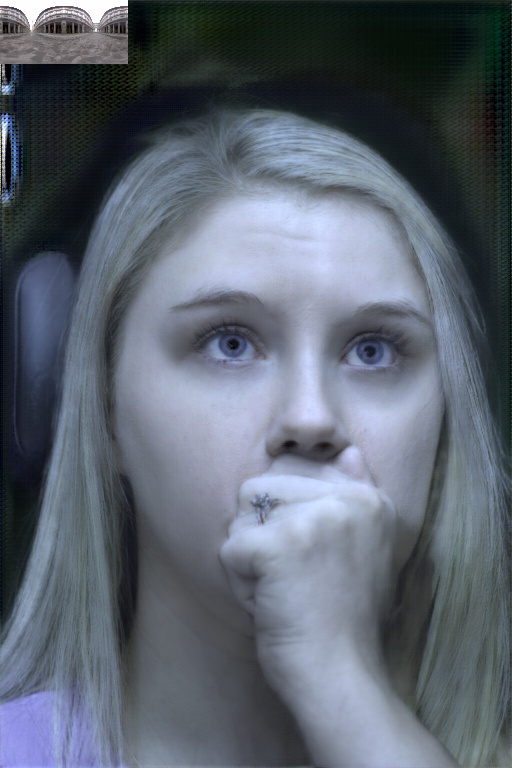}%
\includegraphics[trim=0 11 0 64,clip,
width=\imgsize]{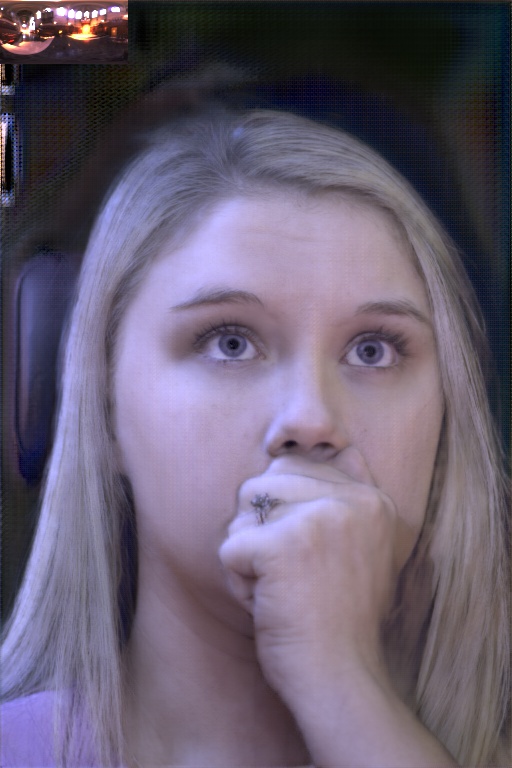}%
\includegraphics[trim=0 11 0 64,clip,
width=\imgsize]{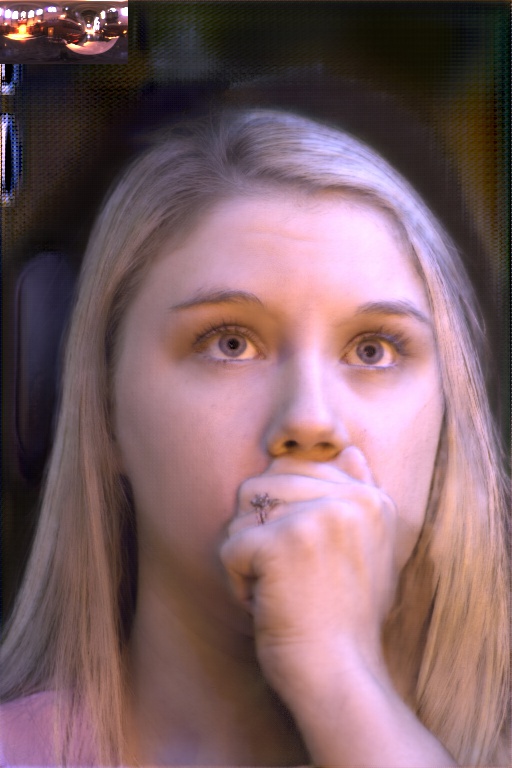}%
\\
\def\subject{152600}
\includegraphics[trim=0 17 0 58,clip,
width=\imgsize]{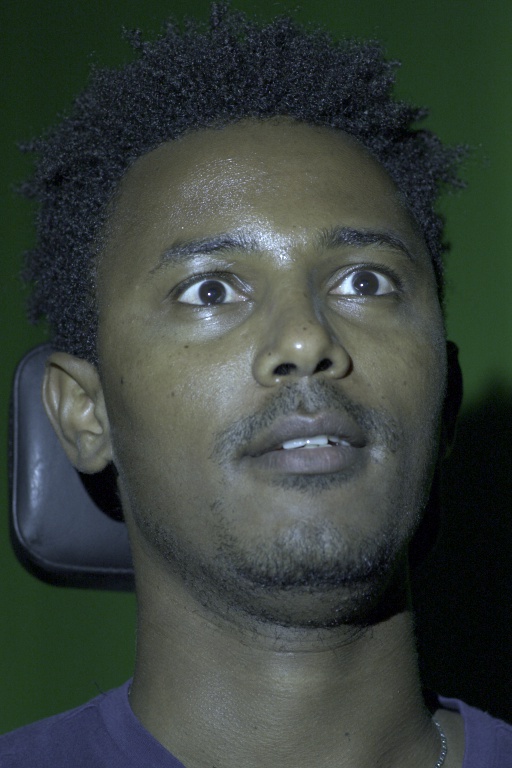}
% \hspace{0.01\linewidth}
\includegraphics[trim=0 17 0 58,clip,
width=\imgsize]{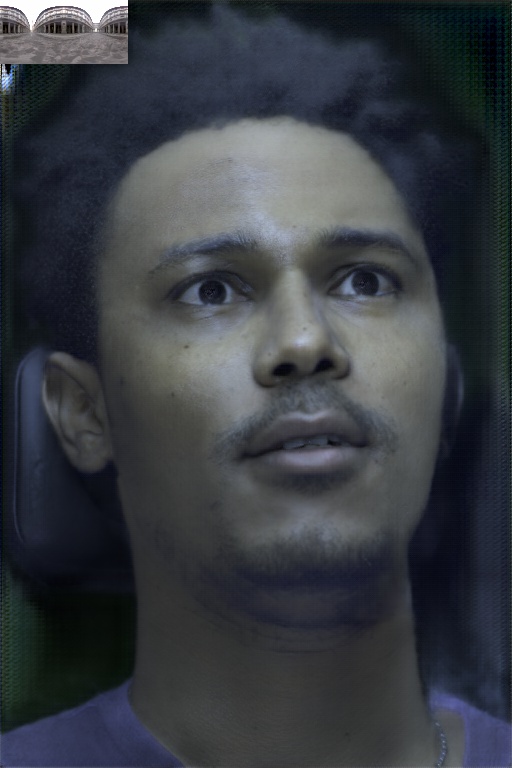}%
\includegraphics[trim=0 17 0 58,clip,
width=\imgsize]{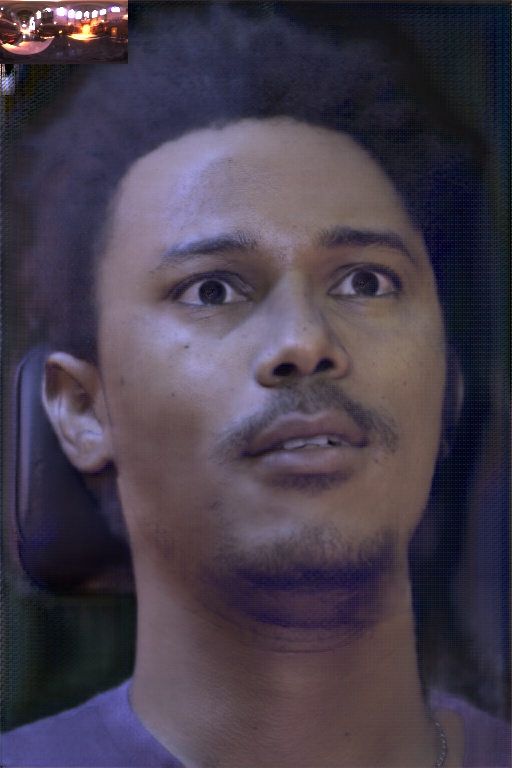}%
\includegraphics[trim=0 17 0 58,clip,
width=\imgsize]{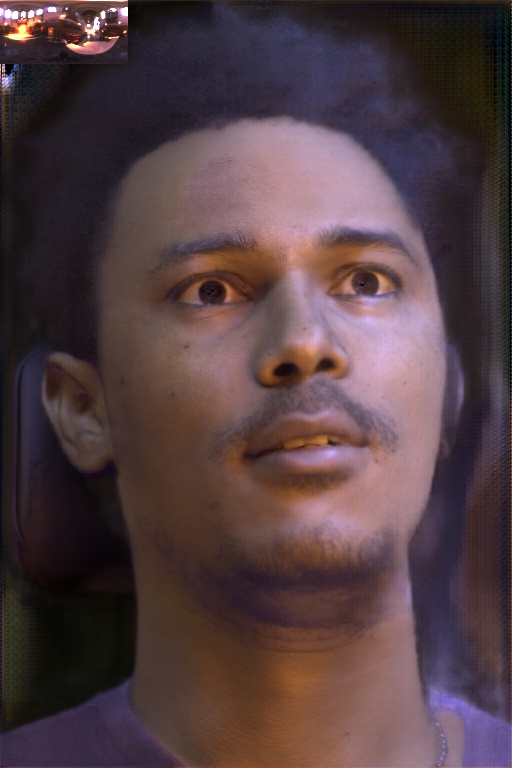}%

\begin{tabular}{@{}C{\imgsize}@{}C{\imgsize}@{}C{\imgsize}@{}C{\imgsize}@{}}
(a) input & (b) env. 1 & (c) env. 2 & (d) env. 3
\end{tabular}

\caption[Relighting with environment maps]{
{\bf Relighting with environment maps.} We consider input images taken under a point light source ({\bf a}) and relight them with respect to $3$ different environment maps ({\bf b-d}).
% (small insets), ordered from cold (second column) to warm (sixth column) dominating color temperatures.
% More results can be found in \cref{fig:experiment:envmap1,fig:experiment:envmap2}.
%All results have been converted from linear to sRGB.
}
\label{fig:experiments:envMap}
\end{figure}

% real data
\subsection{Relighting in the wild}

\begin{figure}%[t]
\setlength{\lineskip}{0pt}
% we currently have (uploaded) index: 0014, 0100, 0170, 0280
\def\lightOne{0170}
\def\lightTwo{0100}
\def\lightThr{0280}
\def\imgType{jpg}
\def\transformed{_transformed}
\centering
%%%%% lights
\includegraphics[width=\imgsize]{figures/lights/miniMugsy/256/0009.png}%
\hfill
\includegraphics[width=\imgsize]{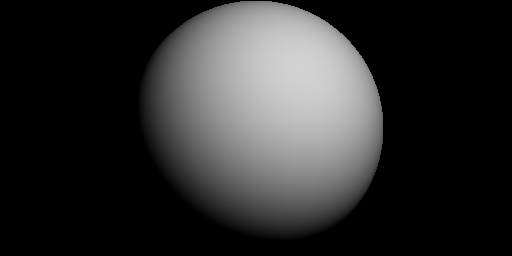}%
\includegraphics[width=\imgsize]{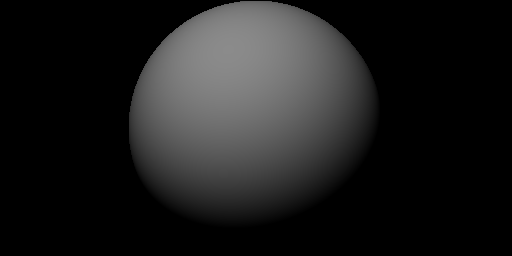}%
\includegraphics[width=\imgsize]{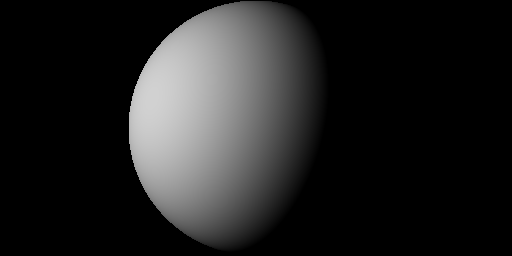}
\\
%%%%% first row
\def\subject{8A2980}
\includegraphics[trim=0 25 0 50,clip,width=\imgsize] {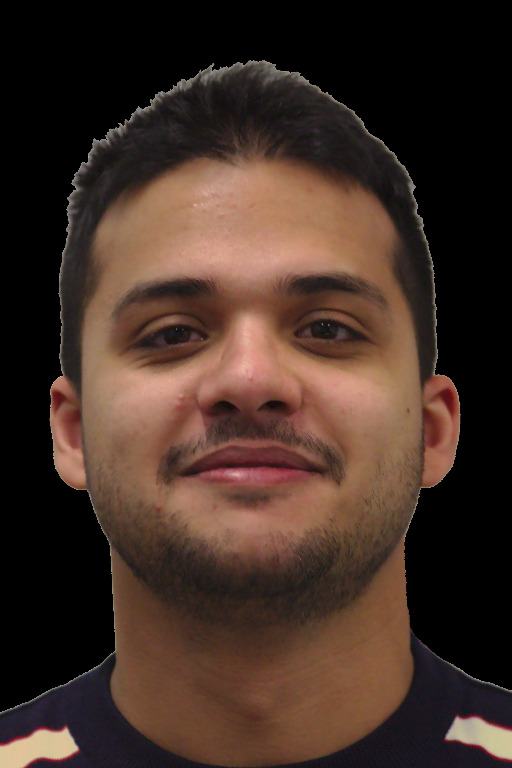}%
\hfill
\includegraphics[trim=0 25 0 50,clip,width=\imgsize] {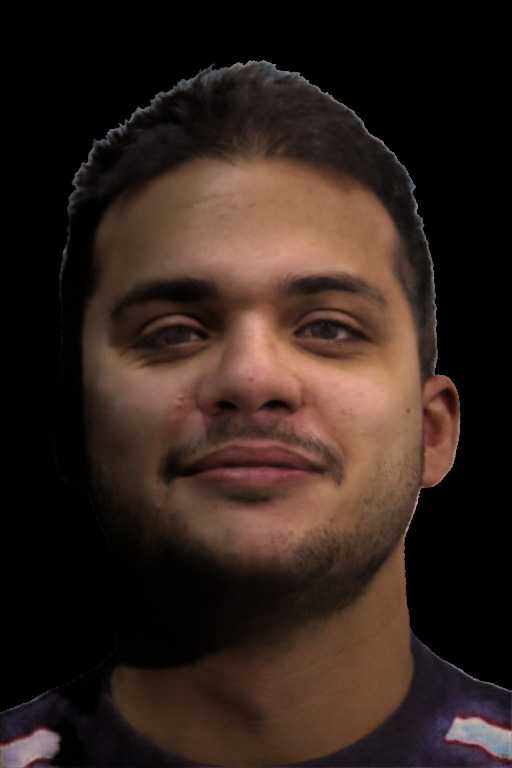}%
\includegraphics[trim=0 25 0 50,clip,width=\imgsize] {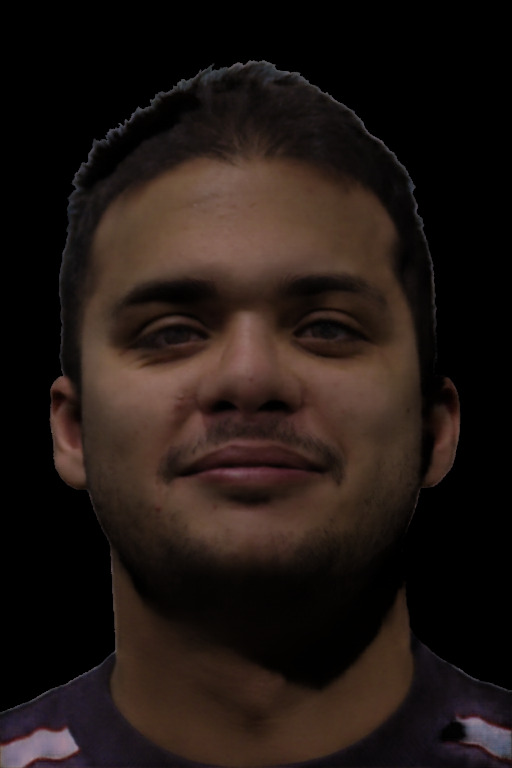}%
\includegraphics[trim=0 25 0 50,clip,width=\imgsize] {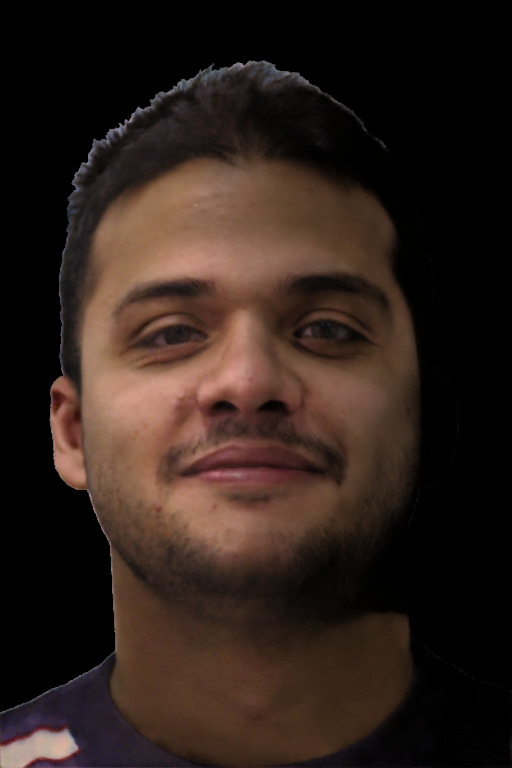}
\\
%%%%% second row
\def\subject{8A2989}
\includegraphics[trim=0 25 0 50,clip,width=\imgsize] {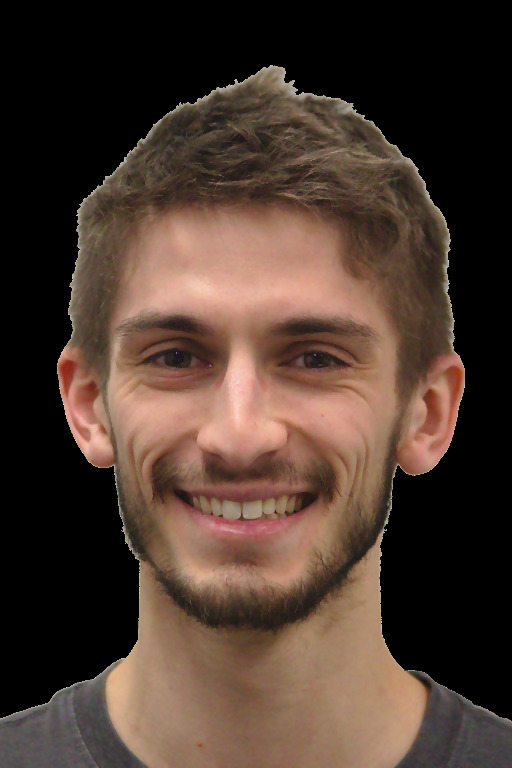}%
\hfill
\includegraphics[trim=0 25 0 50,clip,width=\imgsize] {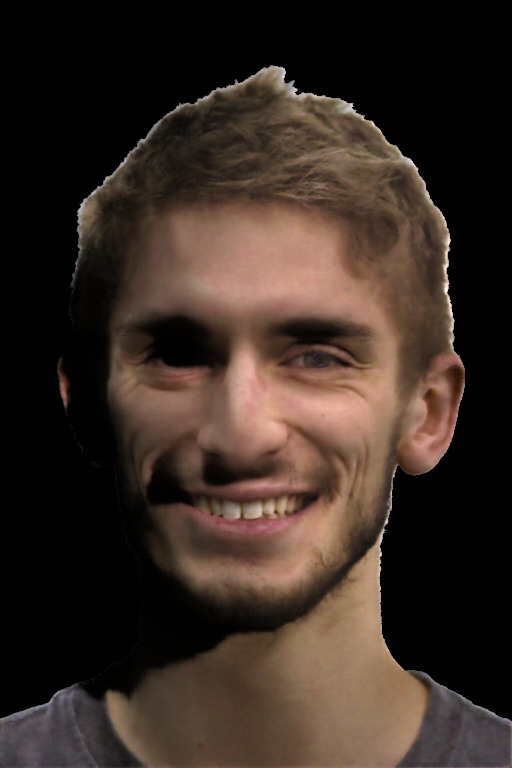}%
\includegraphics[trim=0 25 0 50,clip,width=\imgsize] {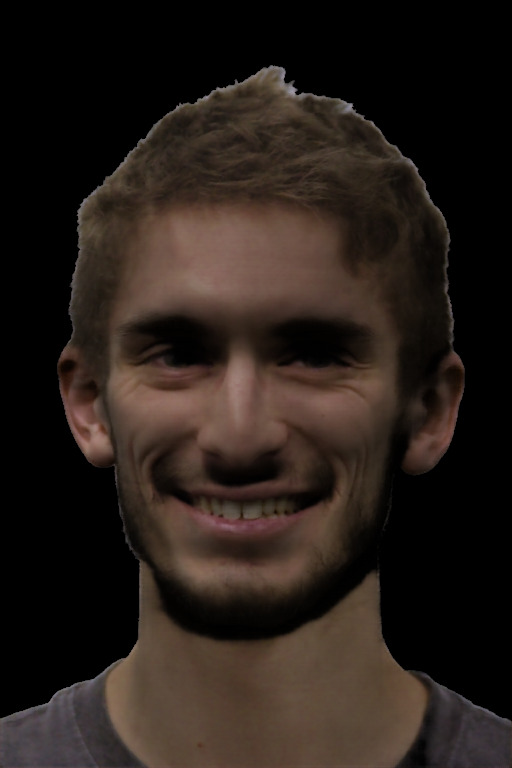}%
\includegraphics[trim=0 25 0 50,clip,width=\imgsize] {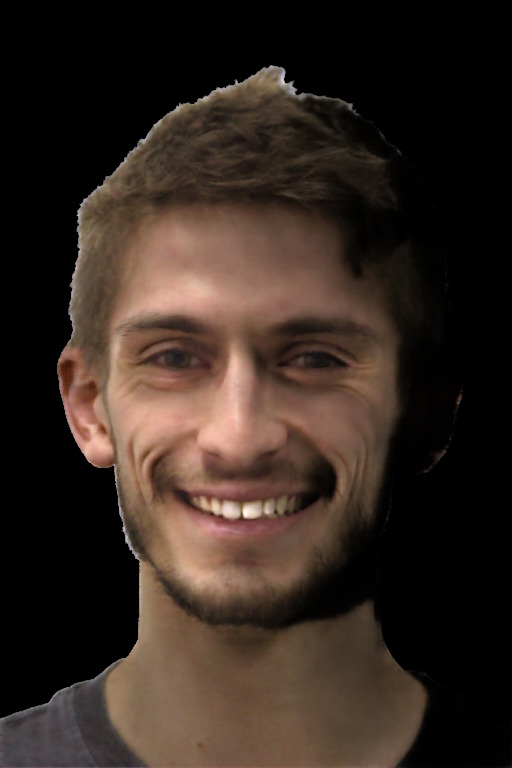}
\\
\begin{tabular}{@{}C{\imgsize}@{}C{\imgsize}@{}C{\imgsize}@{}C{\imgsize}@{}}
(a) input & (b) relit 1 & (c) relit 2 & (d) relit 3
\end{tabular}

\caption{{\bf Relighting in the wild.} We consider portraits not taken in our capture environment ({\bf a}) and relight them with respect to~$3$ different target point lights ({\bf b-d}).} %Point light directions are visualized by rendered spheres at the top.}
\label{fig:real}
\end{figure}

To demonstrate generalization outside the domain of our lab-captured dataset, we conducted experiments using pictures taken in an office environment with a Canon EOS 5D Mark III and visualize results of relighting towards three target lights %$\ell_\text{dst}$
in~\cref{fig:real}.

We emphasize the practical difficulties of relighting those portraits, including unknown discrepancies in the imaging pipeline (camera sensor, illuminant color, image processing etc.) and approximation of the unknown source lighting. Since the portraits are taken under uncontrolled office lighting, this results in images which are diffusely lit by multiple input light sources. Although this violates our model assumption of a single directional light source, we run our model using an input light direction~$\ell_\text{src}$ in the image that would simply light the portrait centrally. To compensate for different illumination colors in the input, we compute the mean of a $51\times 76$ center patch and apply a linear color transform towards our data distribution. %, that we inverse again for visualization.
The background is masked by hand, which could be automated~\cite{wadhwa2018mask}. Note that we can show neither quantitative nor qualitative comparisons to ground truth relit images since they do not exist.

%%%%%%%%%%%%%%%%%%%%%%%%%%%%%%%%%%%%%%%%%%%%%%%%%%%%%%%%%%%%%%%%%%%%%%
% real data in the wild

%!TEX root = 2020CVPR_Relighting.tex
\section{Conclusion}
We propose a method which learns to relight a face with strong directional lighting, accurately reproducing non-diffuse effects such as specularities and hard-cast shadows. 
We introduce a structured relighting architecture with semantic decomposition into, and subsequent re-rendering from intrinsic components. 
On our challenging light-stage dataset with directional light, the integration of an explicit generative rendering process and a non-diffuse neural refinement layer within an end-to-end architecture proved to be superior.
We found that our model tends to produce shadows and albedo maps that are qualitatively closer to reality than all baselines.
A more structured approach also has advantages beyond raw performance, including better interpretability/explainability and the possibility for direct manipulation/extraction of its semantically meaningful intermediate layers for downstream tasks.
A comprehensive study of different losses and evaluation metrics highlighted the benefits of training on a perceptual loss over more traditional choices. We therefore believe our model to be useful in a wide range of face-centric and more general applications in, e.g., augmented reality.

\paragraph{Limitations and future work.}
While our model generally deals well with cast shadows in the input image (see~\cref{fig:massTransfer}), results get worse when there is so little light that the camera mostly returns noise.
Although a crude infilling for those pixels based on context can be learned, an interesting future direction would be to identify these pixels explicitly. A dedicated infilling method, conditioned on the properly relit parts of the image and other intrinsic layers, could be applied to them.
% Also reflections in eyes and glasses are too high frequency for our current model.
To cancel ambient input illumination (as in~\cref{fig:real}), future work could experiment with taking a flash photo and subtracting a second non-flash photo (in linear color space).

% \begin{figure}%[t]
% \centering
% \includegraphics[width=\linewidth]{figures/results/limitations/150910-400005-0005-0016_combined_with_sphere_sRGB.png}
% \caption{In case we have space for a figure for limitations. Considering the complete change in light direction, it even works relatively well.}
% \label{fig:limitations:infilling}
% \end{figure}

\paragraph{Acknowledgments.}
% \section*{Acknowledgments}
We thank Yannick Hold-Geoffroy and Marc-Andr\'{e} Gardner for their valuable feedback,
and Zhixin Shu and Tiancheng Sun for running their code on our dataset.

% \newpage

{\small
\bibliographystyle{ieee_fullname}
\bibliography{relighting}
}

% \newpage
% \appendix
% \input{sec_supplementary.tex}

\end{document}